\documentclass{article}

\PassOptionsToPackage{numbers, compress}{natbib}

\usepackage[final]{neurips_data_2021}

\usepackage[utf8]{inputenc} 
\usepackage[T1]{fontenc}    
\usepackage{hyperref}       
\usepackage{url}            
\usepackage{booktabs}       
\usepackage{amsfonts}       
\usepackage{nicefrac}       
\usepackage{microtype}      
\usepackage{xcolor}         

\usepackage{multirow} 
\usepackage{graphicx}
\usepackage[export]{adjustbox}
\usepackage{float}
\usepackage{epsfig}
\usepackage{graphicx}
\usepackage{amsmath}
\usepackage{amssymb}
\usepackage{caption}
\usepackage{xparse} 
\usepackage{wrapfig} 
\usepackage{tabularx}
\usepackage{subcaption}
\usepackage{fancyhdr}
\usepackage[hang,flushmargin]{footmisc} 

\newcommand{\ques}[1]{\textcolor{black}{#1}}

\usepackage{hyperref}
\hypersetup{
    colorlinks=true,
    urlcolor=magenta,
}

\newcommand{\iconimage}[1]{
\includegraphics[width=12pt, margin=0.5pt 0.1ex 0.5pt 0.1ex, valign=m]{#1}
}

\newlength{\NFwidth}
\NewDocumentCommand{\NFelement}{mmm}{\normalsize\textbf{#1} #2\hfill #3}
\NewDocumentCommand{\NFline}{O{l}m}{\footnotesize\makebox[\NFwidth][#1]{#2}}

\NewDocumentCommand{\NFentry}{sm}{%
  \makebox[.5\NFwidth][l]{\normalsize
    \IfBooleanT{#1}{\makebox[0pt][r]{\textbullet\ }}%
    #2}\ignorespaces}

\newcommand{\NFRULE}{\midrule[5pt]}
\newcommand{\NFRule}{\midrule[2pt]}
\newcommand{\NFrule}{\midrule}

\title{IconQA: A New Benchmark for Abstract Diagram Understanding and Visual Language Reasoning}

\author{%
  Pan Lu$^{1}$, Liang Qiu$^{1}$, Jiaqi Chen$^2$, Tony Xia$^{1}$, Yizhou Zhao$^{1}$,  \\\textbf{ Wei Zhang$^3$, Zhou Yu$^4$, Xiaodan Liang$^2$, Song-Chun Zhu$^{1}$} \\
  $^1$Center for Vision, Cognition, Learning and Autonomy, UCLA \\
  $^2$Sun Yat-sen University, $^3$East China Normal University, 
  $^4$Columbia University
}

\begin{document}

\maketitle

\vspace{-2mm}
\begin{abstract}
Current visual question answering (VQA) tasks mainly consider answering human-annotated questions for natural images. However, aside from natural images, abstract diagrams with semantic richness are still understudied in visual understanding and reasoning research. In this work, we introduce a new challenge of Icon Question Answering (IconQA) with the goal of answering a question in an icon image context. We release IconQA, a large-scale dataset that consists of 107,439 questions and three sub-tasks: \textit{multi-image-choice}, \textit{multi-text-choice}, and \textit{filling-in-the-blank}. The IconQA dataset is inspired by real-world diagram word problems that highlight the importance of abstract diagram understanding and comprehensive cognitive reasoning. Thus, IconQA requires not only perception skills like object recognition and text understanding, but also diverse cognitive reasoning skills, such as geometric reasoning, commonsense reasoning, and arithmetic reasoning. To facilitate potential IconQA models to learn semantic representations for icon images, we further release an icon dataset Icon645 which contains 645,687 colored icons on 377 classes. We conduct extensive user studies and blind experiments and reproduce a wide range of advanced VQA methods to benchmark the IconQA task. Also, we develop a strong IconQA baseline Patch-TRM that applies a pyramid cross-modal Transformer with input diagram embeddings pre-trained on the icon dataset. IconQA and Icon645 are available at \url{https://iconqa.github.io}.

\end{abstract}

\vspace{-2mm}

\section{Introduction}

We are witnessing an exciting development of visual question answering (VQA) research in recent years. The long-standing goal of the VQA task is to exploit systems that can answer natural questions that correspond to visual information. Several datasets have been released to evaluate the systems' visual and textual content understanding abilities \cite{antol2015vqa, zhu2016cvpr, balanced_vqa_v2, johnson2017clevr, hudson2019gqa, wang2020general}. One of the underlying limitations of current VQA datasets is that they are focusing on answering visual questions for natural images. However, aside from natural pictures, abstract diagrams with visual and semantic richness account for a large proportion of the visual world. For instance, it is shown that emojis can express rich human sentiments \cite{kembhavi2016diagram,felbo2017using}, and diagrams like icons can map the physical worlds into symbolic and aesthetic representations \cite{lagunas2019learning, madan2018synthetically, karamatsu2020iconify}. 

\begin{figure}[t]
    \centering 
    \includegraphics[width= 0.85\linewidth]{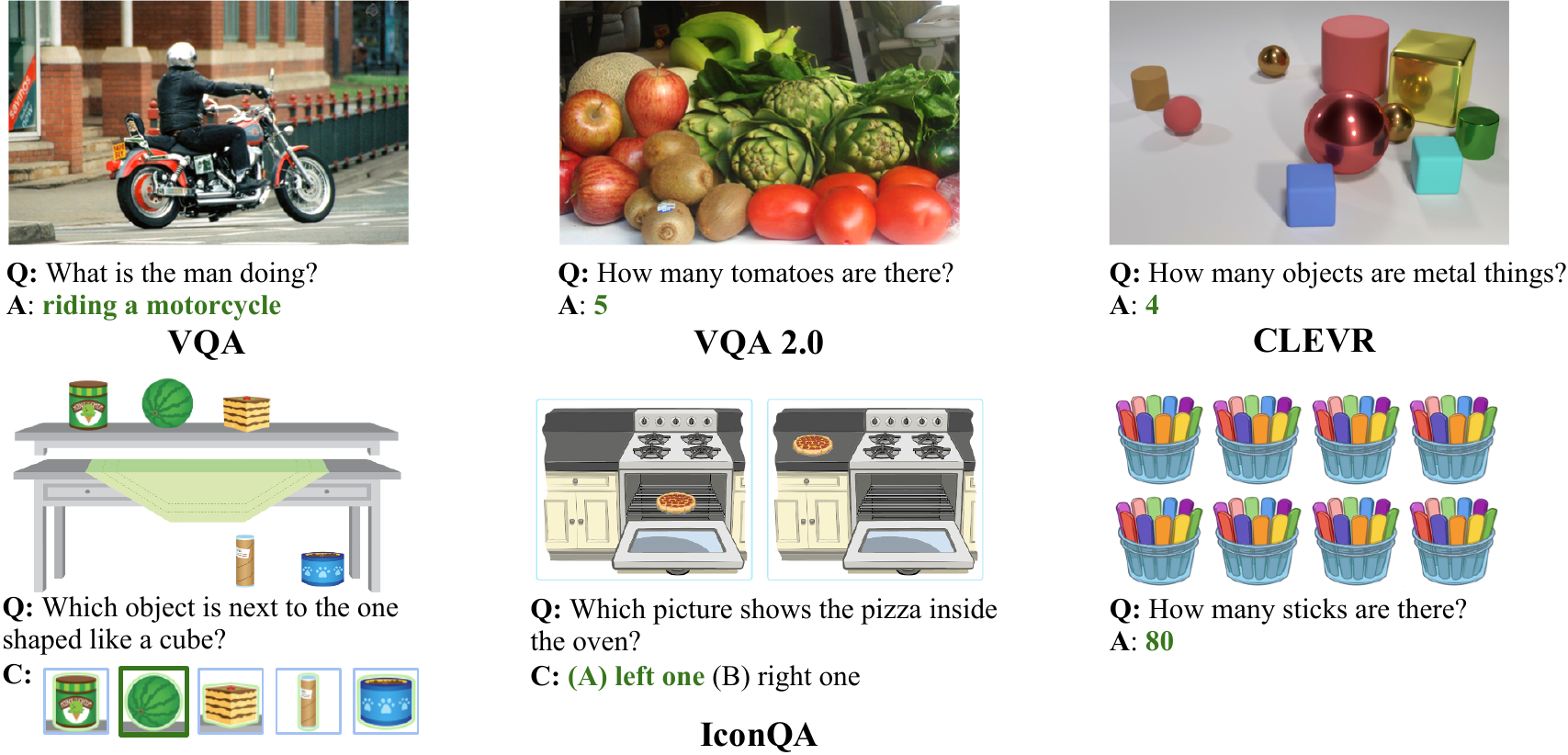}
    \caption{\textbf{Top}: Examples in three popular VQA datasets: VQA \cite{antol2015vqa}, VQA 2.0 \cite{balanced_vqa_v2}, and CLEVR \cite{johnson2017clevr}. \textbf{Bottom}: Examples of three sub-tasks in our IconQA dataset. For answering these icon questions, it requires diagram recognition and text understanding, as well as diverse cognitive reasoning skills.}
    \vspace{-2mm}
    \label{fig1:example}
\end{figure}

Some pioneering works attempt to propose datasets that are capable of answering questions for abstract diagrams. However, these datasets either address domain-specific charts, plots, and illustrations \cite{kembhavi2016diagram,kafle2018dvqa}, or are generated from limited templates \cite{balanced_binary_vqa,johnson2017clevr}. These limitations impede their practical applications in real-world scenarios. For example, in elementary school, abstract diagrams in math world problems are involved with diverse objects and various reasoning skills \cite{karamustafaouglu2011improving}.

To address these shortcomings, we introduce Icon Question Answering (IconQA), a new challenge for \textit{abstract diagram} visual reasoning and question answering. The task, stemming from math word problems for children \cite{martiniello2008language}, exhibits a promising potential to develop education assistants. We name the proposed task as IconQA because the images depict icons, which simplify recognition and allow us to focus on reasoning skills for further research. We release IconQA, a large-scale dataset that contains 107,439 QA pairs and covers three different sub-tasks: \textit{multiple-image-choice}, \textit{multiple-text-choice} and \textit{filling-in-the-blank}. A typical IconQA problem is provided with an icon image and a question, and the answer is in the form of either a short piece of text or a choice from multiple visual or textual choices. Correctly answering IconQA questions needs diverse human intelligence skills. As the examples in Figure \ref{fig1:example} show, IconQA poses new challenges for abstract diagram understanding like recognizing objects and identifying attributes. Besides, it is critical to develop diverse cognitive reasoning skills, including counting objects, comparing attributes, performing arithmetic operations, making logical inferences, completing spatial reasoning, or leveraging external commonsense to answer IconQA questions. More examples from the dataset are shown in Appendix \ref{app_more_examples}.

We use the IconQA dataset to benchmark various VQA approaches in the IconQA task, including four attention-based multimodal pooling methods \cite{Anderson2017up, Kim2018, yu2019mcan, gao2019dynamic} and four Transformer-based pre-trained methods \cite{li2019visualbert, chen2020uniter,wonjae2021an,pmlr-v139-kim21k}, as illustrated in Figure \ref{fig:baselines}. Also, we conduct extensive user studies to evaluate the performance differences between the algorithms and human beings. \textcolor{black}{Three} blind studies show that the IconQA dataset is robust against biased shortcuts when answering icon questions. We further develop a strong baseline called pyramid patch cross-modal Transformer (Patch-TRM), which effectively learns implicit visual and linguistic relationships in IconQA. Patch-TRM parses the diagrams into patch sequences in a spatial pyramid structure and learns a joint embeddings within a multimodal Transformer. Along with the IconQA dataset, we collect an auxiliary icon dataset, Icon645, that features 645,687 colored icons on 377 object classes. The icon dataset is used to pre-train the diagram embedding module in Patch-TRM to enhance abstract diagram understanding. 




Our contributions can be summarized as 1) we propose a new challenge, IconQA, that requires abstract diagram understanding of icon images and diverse visual reasoning skills; 2) we establish two large-scale datasets: IconQA, a question answering dataset in the icon domain, and Icon645, an icon dataset for model pre-training; 3) we benchmark the IconQA dataset extensively via experiments on eight existing methods and develop a strong multimodal Transformer-based baseline.



\section{Related Works}

\textbf{VQA Datasets.} There have been efforts to develop datasets for the visual question answering (VQA) task since the first large-scale benchmark was introduced in \cite{antol2015vqa}. Early released datasets \cite{balanced_vqa_v2,krishna2017visual,singh2019towards,wang2020general} contain natural images and related questions, where understanding the visual and textual contents is essential for question answering. Some recent datasets introduce questions that involve more diverse visual scenes or require external knowledge to answer, which leads to more complex visual and semantic reasoning for question answering. For example, CLEVR \cite{johnson2017clevr} is a synthetic dataset that serves as a diagnostic test for a range of visual reasoning abilities over combinations of three object shapes. However, these datasets are limited to the natural image domain and pay little attention to abstract diagrams, which also have informative semantics and wide applications.




\textbf{Diagram QA Datasets.} To address the need for vision-and-language reasoning for diagrams, several abstract diagram QA datasets have been developed. For example, abstract VQA \cite{antol2015vqa, balanced_binary_vqa} considers the task of answering questions on abstract scenes. Similarly, NLVR \cite{suhr2017corpus}, FigureQA \cite{kahou2017figureqa}, and DVQA \cite{kafle2018dvqa} feature diagrams that are generated with several figure types or question templates. However, either diagrams or questions in these datasets are generated from limited templates, leading to the existence of unintended visual or linguistic shortcuts for question answering. Some more works have proposed datasets of middle school math or science problems in more practical and complex scenarios \cite{seo2015solving, kembhavi2017you, sachan2017textbooks,sachan2018learning,lu2021inter}.  A central limitation of the subject QA datasets is that they require complex domain-specific knowledge, which makes disentangling visual reasoning and domain knowledge difficult. Herein, we address these limitations by introducing the IconQA dataset, where only elementary commonsense is required. Through IconQA, we aim to provide a new benchmark for abstract scene understanding and learning different visual reasoning skills in \textit{real-world} scenarios.


\textbf{VQA Methods.} Early VQA approaches usually combine multi-modal inputs by applying attention mechanisms over image regions or question words \cite{Kim2018,lu2018co,lu2018rvqa,gao2018question,yu2019mcan,gao2019dynamic}. Inspired by the semantic nature of VQA images, a line of approaches adopt object proposals from pre-trained object detectors and learn their semantic relationships \cite{Kim2018,yu2019mcan,gao2019dynamic}. As Transformers achieve excellent performance on vision tasks, pioneering works have attempted to use pre-trained models to learn visual representations for natural images in the VQA task \cite{lu2019vilbert, li2019visualbert,chen2020uniter,pmlr-v139-kim21k} and achieve significant improvements. However, current VQA models are not capable of extracting meaningful visual representations from abstract diagrams, as they require image embeddings or object proposals learned from natural images. Instead, we develop a strong baseline that feeds spatial patch sequences into a Transformer encoder that is powered by the embedding module pre-trained on our Icon645 dataset. 






\section{The IconQA Dataset}
\label{sec:iconqa}

The IconQA dataset provides diverse questions that require abstract diagram recognition, comprehensive visual reasoning skills, and basic commonsense knowledge. IconQA consists of 107,439 questions split across three different sub-tasks. To the best of our knowledge, IconQA is the largest VQA dataset that focuses on real-world problems with icon images while \textcolor{black}{involving multiple human intelligence reasoning abilities} (see Table \ref{table:dataset}). 

\subsection{Data Collection}

We aim to collect icon-based question answering pairs that \textcolor{black}{involve multiple reasoning skills}, such as visual reasoning and commonsense reasoning. To construct the IconQA dataset, which stems from real-world math word problems, we search for open-source math textbooks with rich icon images and diverse topics. Of those, we choose \textit{IXL Math Learning} which compiles popular textbooks aligned to California Common Core Content Standards\footnote{\url{https://www.ixl.com/standards/california/math}}. We ask well-trained crowd workers to collect problems that cover content from pre-K to third grade, as these problems usually contain abstract images and involve little to none complex domain knowledge. With the driven interest of visual reasoning over abstract images, we filter out the questions that do not accompany icon images or only have images in black and white. Redundant or repetitive data instances are also removed. Question choices are randomly shuffled to ensure a balanced answer distribution. See Appendix \ref{app_iconqa_data} for full details of the dataset collection and usage.


\subsection{Data Analysis}
\label{sec:data_analysis}

Finally, we collect 107,439 IconQA data instances, where each data point contains a colored icon image, a natural language question, optional image or text choices, as well as a correct answer. The IconQA dataset consists of 107,439 questions and is divided into train, validation, and test splits with a ratio of 6:2:2, as shown in Table \ref{table:iconqa_splits}. The dataset consists of three sub-tasks: \textit{multi-image-choice}, \textit{multi-text-choice}, and \textit{filling-in-the-blank}. The \textit{multi-image-choice} sub-task is defined as choosing the correct image from a list of image candidates based on a given diagram and its corresponding question. Similarly, the \textit{multi-text-choice} sub-task is defined as a multiple choice question with 2-5 text choices and an abstract diagram. The \textit{filling-in-the-blank} sub-task is similar to the common VQA task, requiring a brief text answer for each question, except in IconQA, the images are icon images instead of natural images.





\begin{table}[t]
    \begin{minipage}{0.48\linewidth}
        \centering
        \scriptsize
        \renewcommand\tabcolsep{5.0pt}
        \begin{tabular}{lc|ccc}
            \toprule	
            \textbf{Tasks} & \text{All} & \text{Train}  & \text{Val} & \text{Test}  \\ 
            \midrule	
            \textit{Multi-image-choice} & 57,672 & 34,603 & 11,535 & 11,535 \\
            \textit{Multi-text-choice} & 31,578 & 18,946 & 6,316 & 6,316 \\ 
            \textit{Filling-in-the-blank} & 18,189 & 10,913 & 3,638 & 3,638 \\
            \midrule
            All & 107,439 & 64,462 & 21,489 & 21,489 \\
            \bottomrule	
        \end{tabular}
        \vspace{1mm}
        \caption{Statistics for the IconQA dataset.}
        \label{table:iconqa_splits}
    \end{minipage}
    \hfill
	\begin{minipage}{0.48\linewidth}  
        \centering
        \scriptsize
        \renewcommand\tabcolsep{5.0pt}
        \begin{tabular}{lc|ccc}
            \toprule	
            \textbf{Task} & \text{Avg.} & \text{1 skill} & \text{2 skills} & \text{3 skills} \\ 
            \midrule
            \textit{Multi-image-choice}   & 1.51 & 55.78\% & 37.44\% & 6.77\% \\
            \textit{Multi-text-choice}    & 1.73 & 33.21\% & 60.14\% & 6.65\% \\ 
            \textit{Filling-in-the-blank} & 1.81 & 28.30\% & 62.43\% & 9.25\% \\
            \midrule	
            All                  & 1.63 & 44.50\% & 48.34\% & 7.16\% \\
            \bottomrule	
        \end{tabular}
        \vspace{1mm}
        \caption{Skill numbers for questions in IconQA.}
        \label{table:skill_num}
    \end{minipage}
\end{table}

\begin{figure*}[t]
    \begin{minipage}{0.27\linewidth}  
        \centering
        \includegraphics[width= 1.\linewidth]{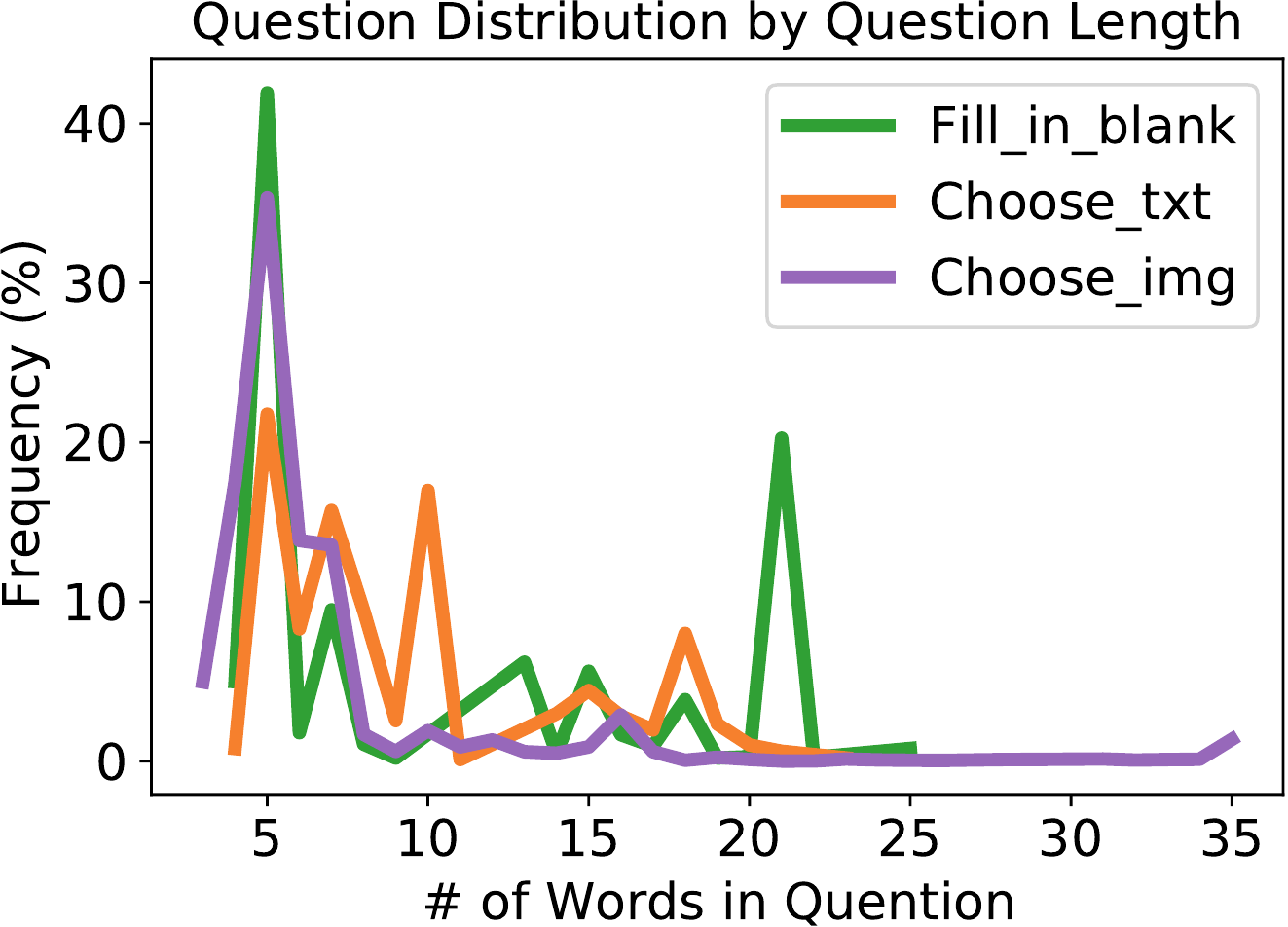}
        \vspace{-5mm}
        \subcaption*{(a)}
	\end{minipage}
    \hfill
    \begin{minipage}{0.71\linewidth}  
        \centering
        \includegraphics[width= 1.\linewidth]{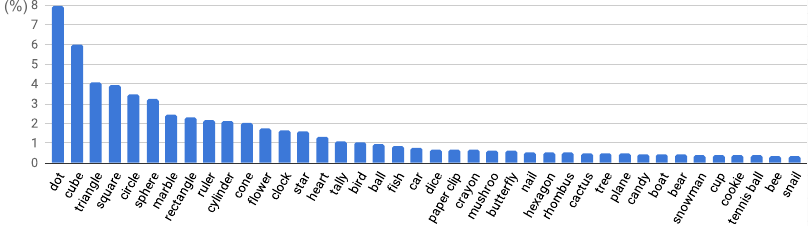}
        \vspace{-5mm}
        \subcaption*{(b)}
	\end{minipage}
	  \vspace{-3mm}
      \caption{\text{(a)} Question statistics based on number of words. \text{(b)} Top 40 icons mentioned in the IconQA question texts and their appearance percentage. These icons cover various types of real-world objects.}
    \label{fig:question}
\end{figure*}

\begin{figure}[ht]
    \vspace{-1mm}
    \begin{minipage}{0.39\linewidth}  
        \centering 
        \includegraphics[width= 0.93\linewidth]{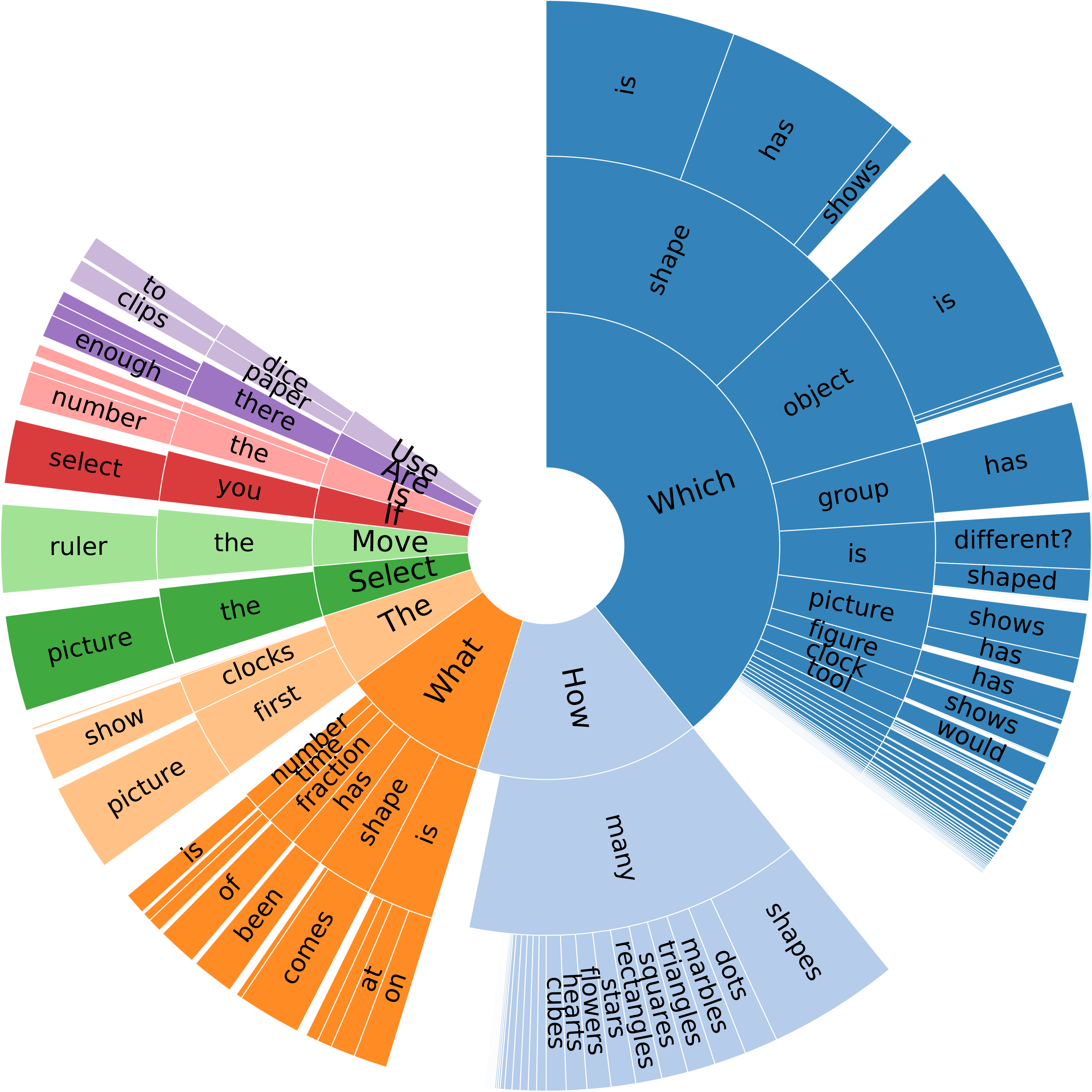}
        \caption{Question types in IconQA.}
        \label{fig:question_pie}
	\end{minipage}
	\hfill
	\begin{minipage}{0.60\linewidth}  
        \centering 
        \includegraphics[width= 0.96\linewidth]{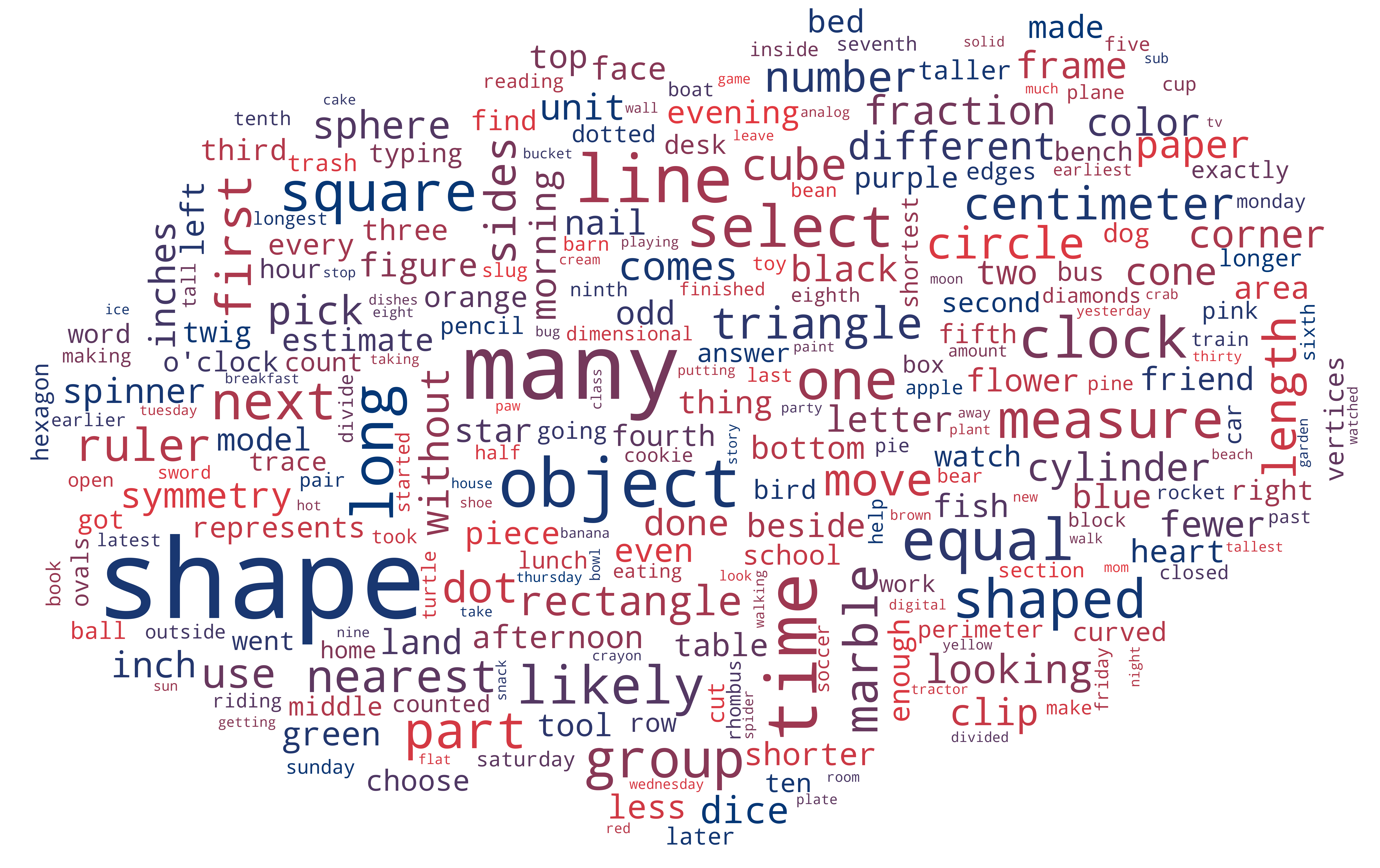}
        \caption{Word cloud of the question text in IconQA.}
        \label{fig:wordcloud}
    \end{minipage}
    \vspace{-1mm}
\end{figure}

\textbf{Questions.} Figure \ref{fig:question}  (a) illustrates the distribution of question lengths of each sub-task in the IconQA dataset. For simplicity, all questions longer than 35 words are counted as having 35 words. Questions in the \textit{multi-text-choice} sub-task distribute more evenly, while for \textit{multi-img-choice}, there is a long-tail distribution due to the complexity of textual scenarios. We find that some icon objects are frequently mentioned in the questions. In Figure \ref{fig:question} (b), the frequencies of the 40 most frequently mentioned icons are shown. These icon entities cover different daily-life objects such as animals, plants, shapes, food, etc. We cluster question sentences into different types based on frequent trigram prefixes starting the sentences. The distribution of questions is visualized in Figure \ref{fig:question_pie}. Importantly, the diversity in the question distribution implies the requirement of high-level understanding of textual and visual contents in IconQA. Figure \ref{fig:wordcloud} shows the word cloud of the question text in IconQA after eliminating the stop words. The most frequent words: \textit{shape}, \textit{many}, and \textit{object} indicate that answering IconQA questions requires the model to identify a variety of geometric shapes and icon objects. Inspired by this, learning informative representations for icon images plays an important role in visual reasoning for the IconQA task. 

\begin{figure}[ht] 
  \vspace{-1mm}
  \begin{minipage}[b]{0.48\textwidth} 
    \centering
    \fontsize{7.5pt}{\baselineskip}\selectfont
    \renewcommand\tabcolsep{1.0pt}
    \renewcommand\arraystretch{0.75}
    \begin{tabular}{ll}
        \toprule
        \textbf{Skill types} & \text{Description} \\
        \midrule
        Geometry & Identify shapes, symmetry, transformations\\
        Counting & Count objects, shapes \\
        Comparing & Compare object attributes \\
        Spatial & Identify spatial positions and relations \\
        Scene & Understand abstract scenes \\
        Pattern & Identify next and different patterns \\
        Time & Identify time of clocks, events \\
        Fraction & Perform fraction operations \\
        Estimation & Estimate lengths, large numbers  \\
        Algebra & Perform algebraic operations \\
        Measurement & Measure widths, lengths, heights \\
        Commonsense & Apply external knowledge \\
        Probability & Perform probability and statistics operations \\ \bottomrule
    \end{tabular}
    \captionof{table}{Definition of reasoning skill types.}
    \label{table:skills}
  \end{minipage} 
  \begin{minipage}[b]{0.48\textwidth} 
    \centering 
    \includegraphics[width=0.85\textwidth]{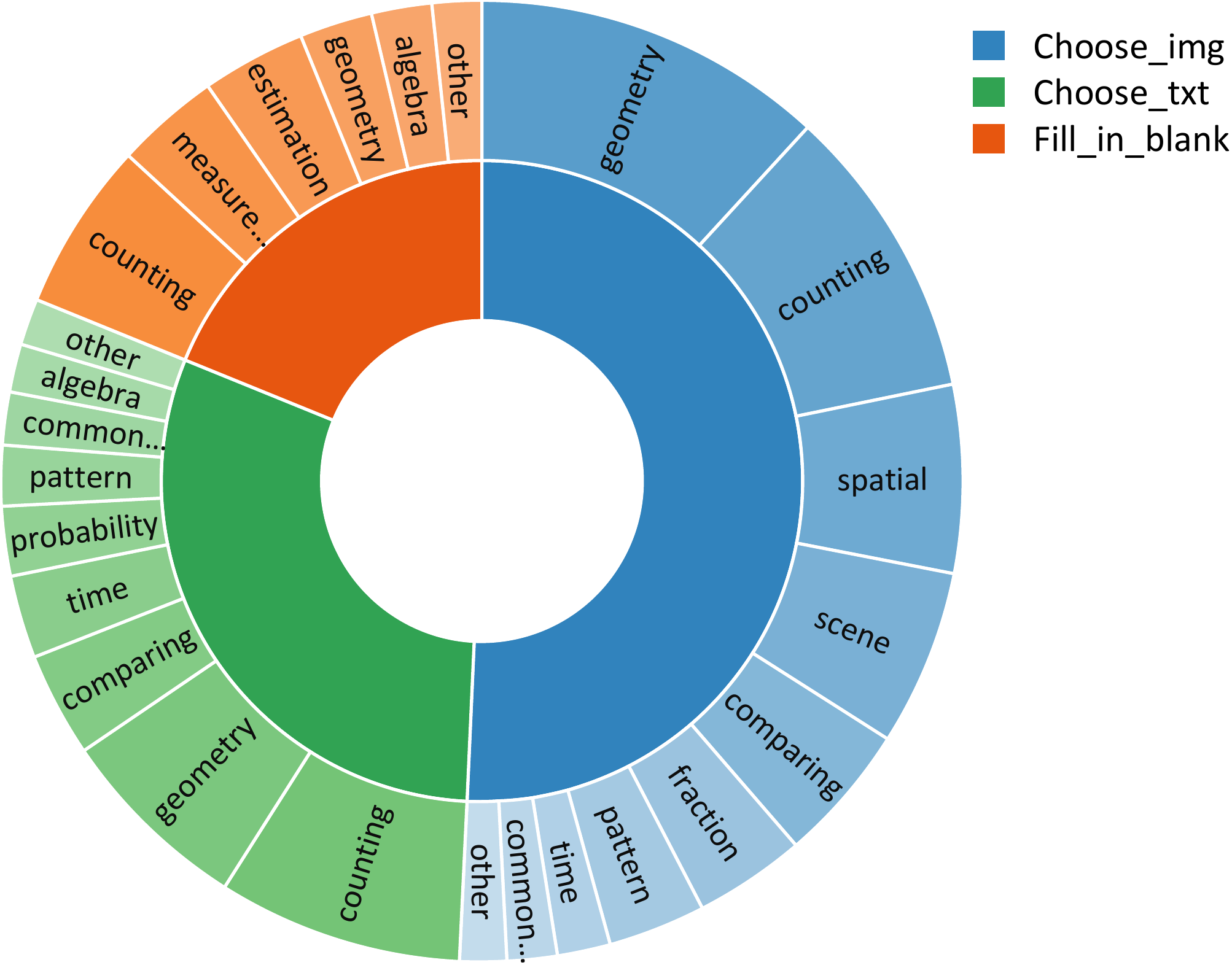} 
    \caption{Skill distribution in IconQA questions.} 
    \label{fig:skill_dist} 
  \end{minipage}
  \vspace{-2mm}
\end{figure}

\textbf{Skill Categories.} Our IconQA dataset contains questions of multiple different cognitive reasoning and arithmetic reasoning types that can be grouped into 13 categories, shown in Table \ref{table:skills}. We annotate each question in IconQA with its corresponding skill types based on the tags provided by the original problem sources. Figure \ref{fig:skill_dist} shows the distributions of questions related to each skill. For instance, to answer 13.8\% of the questions in IconQA, the model has to be capable of \textit{comparing} object attributes. Additionally, each question can be related to up to three skills out of these 13 categories, and on average, a question requires 1.63 skills. The detailed statistics are demonstrated in Table \ref{table:skill_num}. In general, the \textit{filling-in-the-blank} sub-task consists of questions that require the most number of skills, averaging 1.81 skills per question. 9.25\% of the \textit{filling-in-the-blank} questions require 3 skills. As the examples from IconQA shown in Figure \ref{fig1:example}, the first and second questions require the skills  of \textit{scene} understanding and \textit{spatial} reasoning. The third example asks how many sticks exist in the diagram, requiring the basic ability of \textit{counting} and basic \textit{algebra} operations. As stated before, the IconQA dataset requires a wide range of skills for a model to perform well on IconQA. 


\textbf{Comparisons to Other Datasets.} We compare our IconQA dataset with two datasets on natural images and five datasets on abstract diagrams in Table \ref{table:dataset}. To summarize, IconQA is different from these datasets in various aspects. Unlike natural images (VQA \cite{antol2015vqa}, CLEVR \cite{johnson2017clevr}) or abstract diagrams like scenes, charts, plots, and illustrations (VQA-Abstract \cite{antol2015vqa}, DVQA \cite{kafle2018dvqa}, NLVR \cite{suhr2017corpus}, AI2D \cite{kembhavi2016diagram}, Geometry3K \cite{lu2021inter}), IconQA features icon images and covers the largest object set of 388 classes. As questions in IconQA stem from real-world math problems and they may describe complex problem scenarios, IconQA has the longest question length among all related datasets. Furthermore, IconQA requires both commonsense and arithmetic reasoning due to its origin from real-world problems. Lastly, IconQA contains more QA task types including answering questions with image choices.

\begin{table*}[h]
\centering
\scriptsize
\renewcommand\tabcolsep{2pt}
\renewcommand{\arraystretch}{1.0}
\begin{tabular}{{l}*{13}{c}}
    \toprule	
    & \text{\#QA} & \text{\#Image} & \text{AvgQ} & \text{MaxQ} & \text{Image Type} & \text{Source} & \text{\#Object} & \text{\#Task} & \text{VisualAns} & \text{CommonSen} & \text{Arithmetic} \\ 
    \midrule
    \text{VQA} \cite{antol2015vqa} & 614,163 & 204,721 & 6.1 & 23 & 
    Natural & Annotated &  - & 2 &  & \checkmark & \\
    \text{CLEVR} \cite{johnson2017clevr} & 999,968 & 100,000 & 18.4 & 43 & 
    Natural & Generated & 3 & 1 &  &  & \\
    \text{VQA-Abstract} \cite{antol2015vqa} &  150,000 & 50,000 & 6.0 & 21 & 
    Scene & Annotated & 131 & 2 &  &  & \\
    \text{DVQA} \cite{kafle2018dvqa} & 2,325,316 & 300,000 & 10.3 & 23 & 
    Bar chart & Generated & - & 1 &  &  & \checkmark \\
    \text{NLVR} \cite{suhr2017corpus} & 92,244  &92,244  & 11.2 & 25 & 
    Scatter plot & Generated & 3 & 1 &  &  \\
    \text{Geometry3K} \cite{lu2021inter} & 3,002  &2,342  & 10.1 & 46 & 
    Diagram & Real-world & 4 & 1 &  & & \checkmark  \\
    \text{AI2D} \cite{kembhavi2016diagram} & 4,563 & 4,903 & 9.8 & 64 & 
    Illustration & Real-world & - & 1 &  & \checkmark \\
    \textbf{IconQA} (Ours) & 107,439 & 96,817 & 8.4 & 73 & 
    Icon image & Real-world & 388 & 3 & \checkmark & \checkmark & \checkmark \\
    \bottomrule
\end{tabular}
\caption{Statistics for the IconQA dataset and comparisons with existing datasets. Dataset source: \textit{real-world} datasets refer to those that are collected from textbooks or online resources, not manually \textit{annotated} or automatically \textit{generated}.}
\label{table:dataset}
\end{table*}


\subsection{Impact and Ethics}
\label{sec:impact}

\textbf{Impact \& Usage.} IconQA is useful for not only follow-up research projects but also real-world applications (e.g.  K-6 education applications like tutoring assistants). Moreover, visual recognition in the abstract domain is essential to general AI agents, but rarely investigated in the community, posing new challenges in abstract and symbolic visual reasoning -- a natural ability of human.

\textbf{Social Ethics.} 
Unlike VQA datasets in the natural image domain, IconQA is completely built upon abstract icon images. Therefore, it is less likely to be used in surveillance systems that might infringe on people's privacy. Moreover, due to the abstract nature of the dataset, IconQA does not contain any sensitive personal information such as gender and race, nor does it contain data that might exacerbate biases against under-represented communities. Therefore, after careful examinations of our dataset, we think the dataset is unlikely to be used to harm people directly. 

\section{The Icon645 Dataset}
\label{sec:Icon645}
As discussed in Section \ref{sec:data_analysis}, IconQA questions are accompanied by abstract diagrams that cover a wide range of icon objects. Using existing backbone networks to extract image representations for these icon images is inadequate, as most of these networks are pre-trained on natural images. To overcome the limitation, we develop a new large-scale icon dataset for pre-training existing vision backbone networks. We use the collected icon data to pre-train the current backbone networks, which can be applied to extract diagram representations in IconQA.


We retrieve the 388 icon classes mentioned in the question texts from Flaticon\footnote{Flaticon: \url{https://www.flaticon.com/}}, the largest database of free vector icons. After removing 11 classes that can't be retrieved, we construct an icon dataset containing 377 classes, called Icon645. As summarized in Table \ref{table:icon} (Appendix), the Icon645 dataset includes 645,687 colored icons with a minimum size of 64 by 64 and a maximum size of 256 by 256. Examples in Table \ref{table:icon_examples} show that our collected icons include a wide variety of colors, formats and styles. On top of pre-training encoders, the large-scale icon data could also contribute to future research on abstract aesthetics and symbolic visual understanding. In this work, we use the icon data to pre-train backbone networks on the icon classification task in order to extract semantic representations from abstract diagrams in IconQA. See Appendix \ref{app_icon645} for the details of data collection and analysis.

\begin{table}[t]
\centering
\scriptsize
\begin{tabular}{lc|lc}
    \toprule	
    \textbf{Icons} & \text{Examples} & \textbf{Icons} & \text{Examples}  \\ 
    \midrule	
    Bed & 
    \iconimage{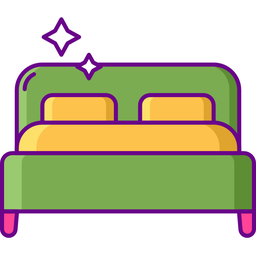} 
    \iconimage{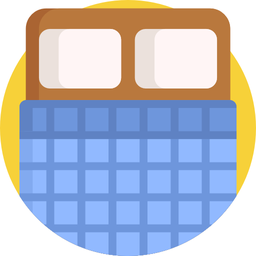} 
    \iconimage{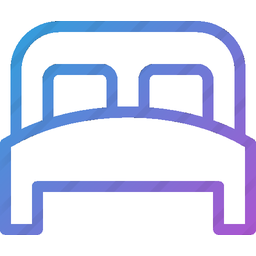} 
    \iconimage{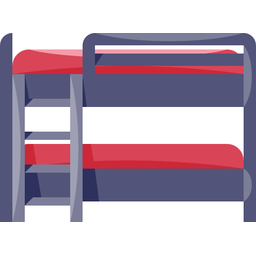}
    \iconimage{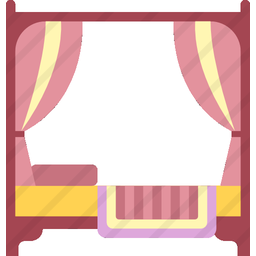}
    \iconimage{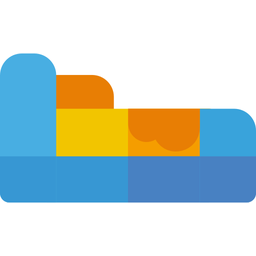}  
    \iconimage{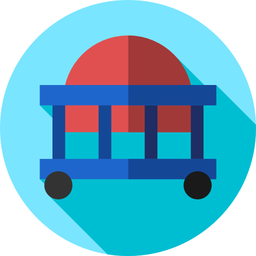} & 
    Bucket & 
    \iconimage{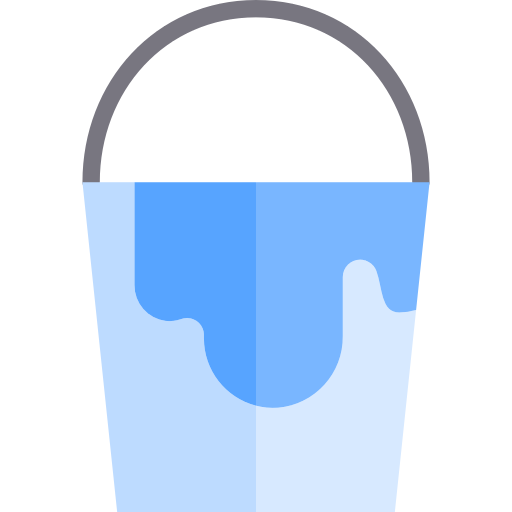} 
    \iconimage{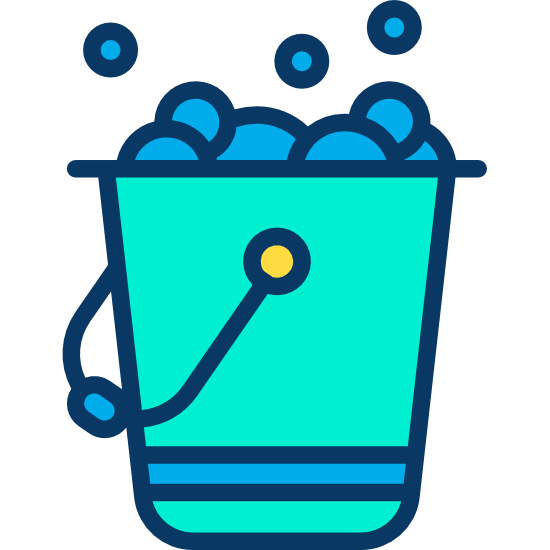} 
    \iconimage{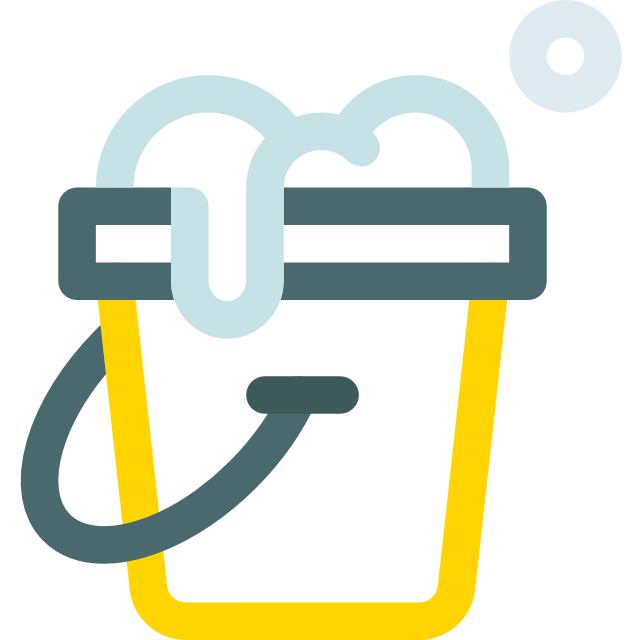} 
    \iconimage{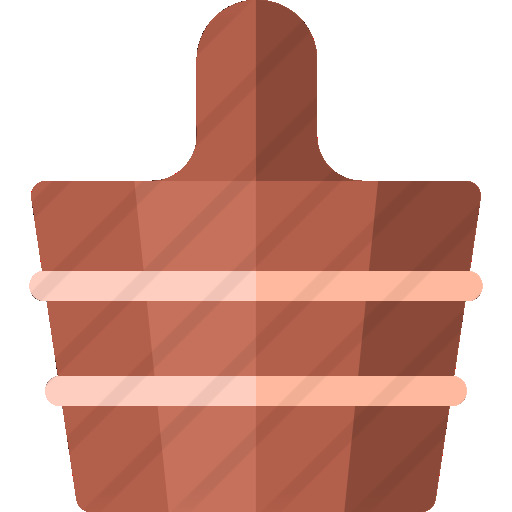}
    \iconimage{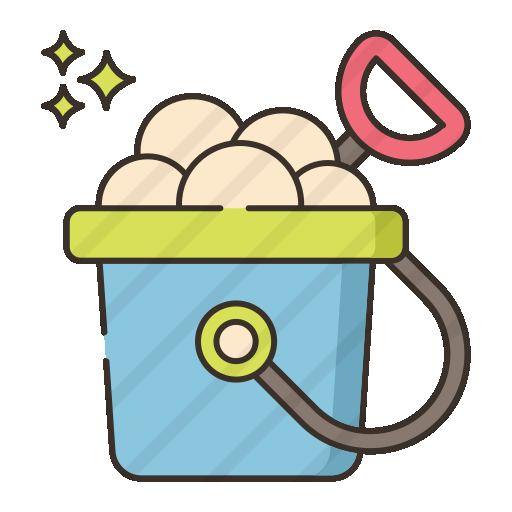}
    \iconimage{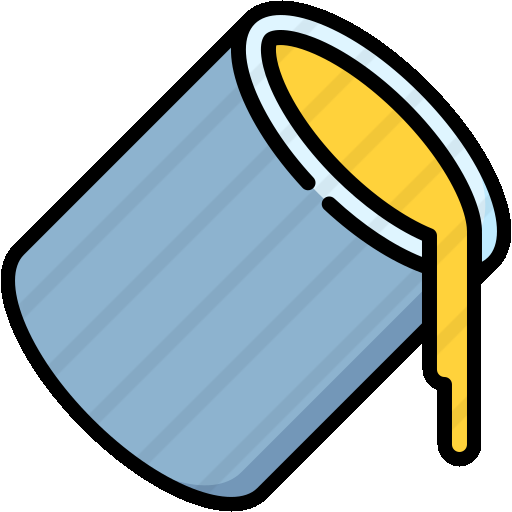}  
    \iconimage{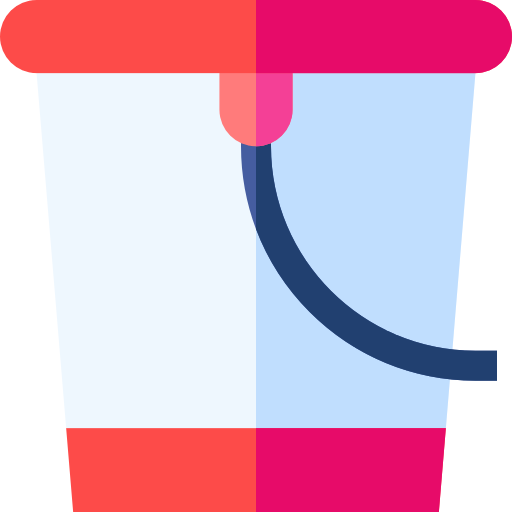} \\
    Cake & 
    \iconimage{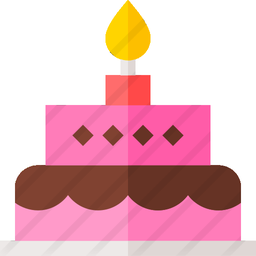}  
    \iconimage{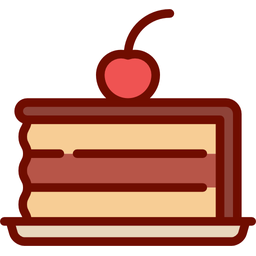} 
    \iconimage{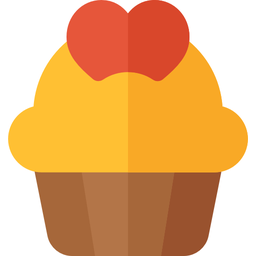}  
    \iconimage{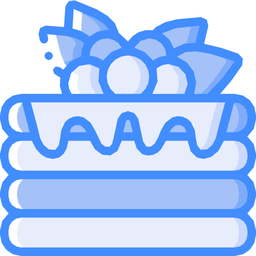}
    \iconimage{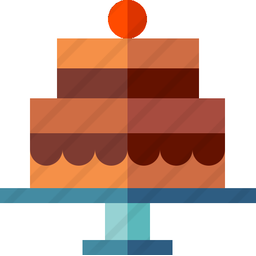}
    \iconimage{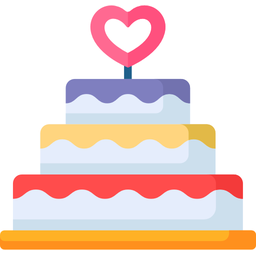}
    \iconimage{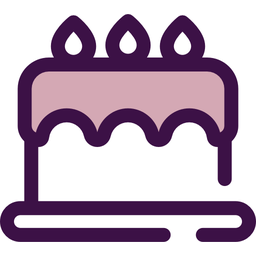} & 
    Car & 
    \iconimage{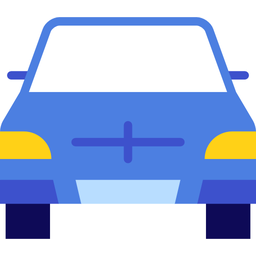} 
    \iconimage{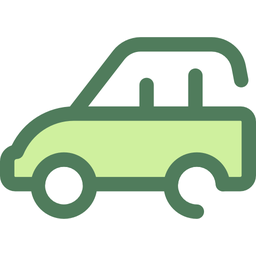} 
    \iconimage{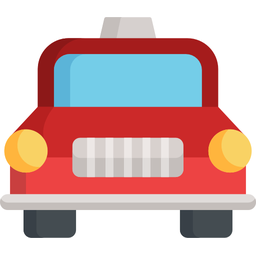} 
    \iconimage{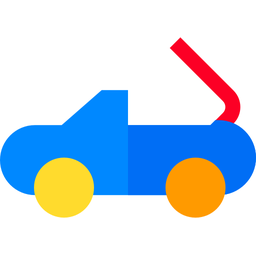}
    \iconimage{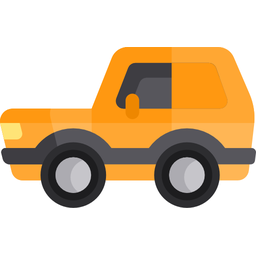}
    \iconimage{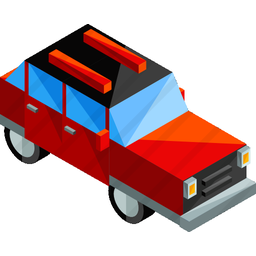}  
    \iconimage{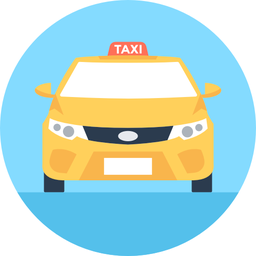}  \\ 
    Castle & 
    \iconimage{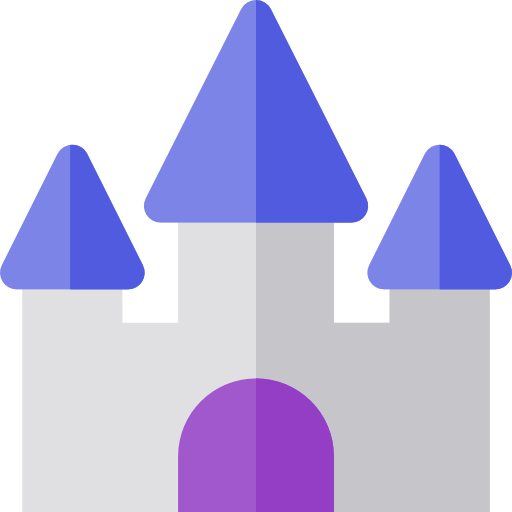} 
    \iconimage{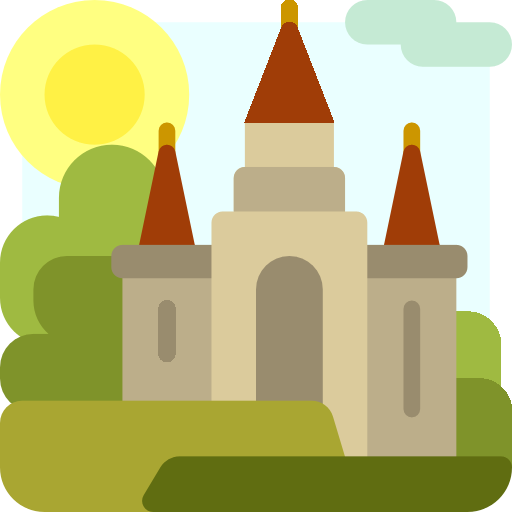} 
    \iconimage{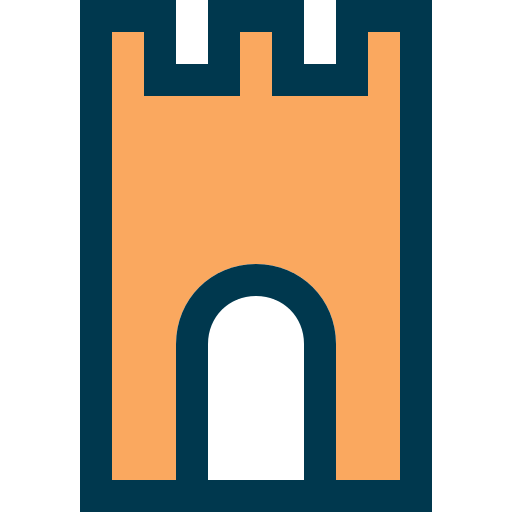} 
    \iconimage{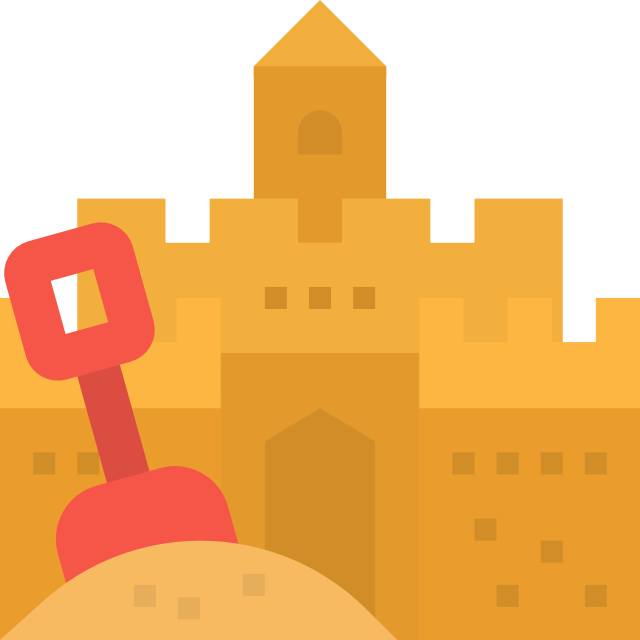}
    \iconimage{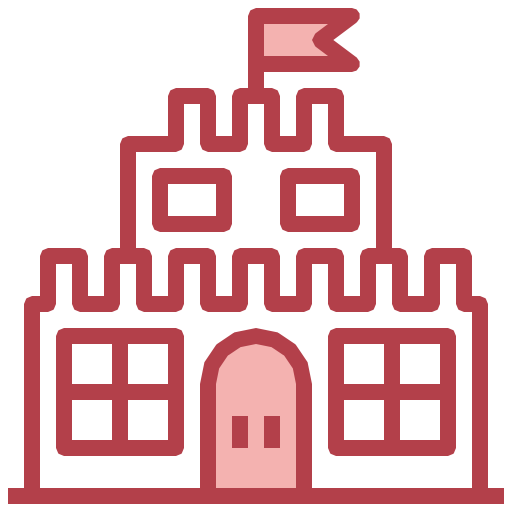}
    \iconimage{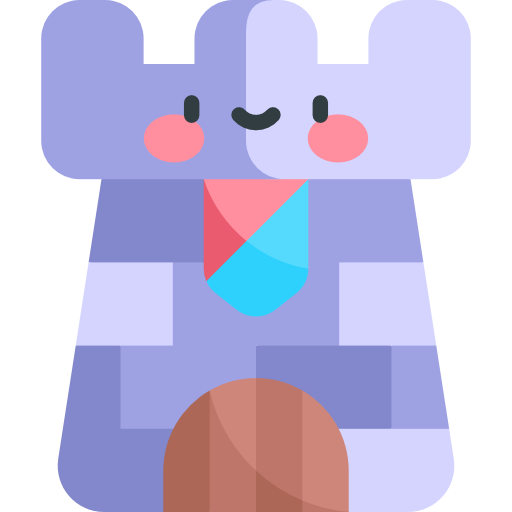}  
    \iconimage{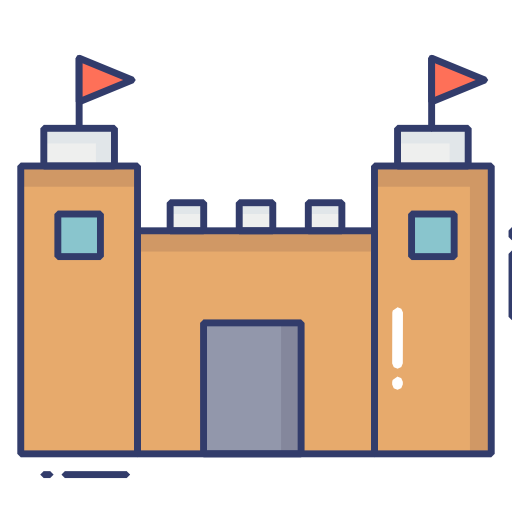} & 
    Dog & 
    \iconimage{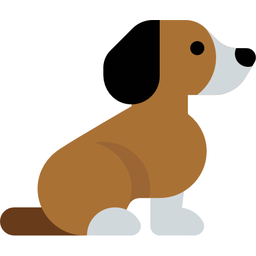}  
    \iconimage{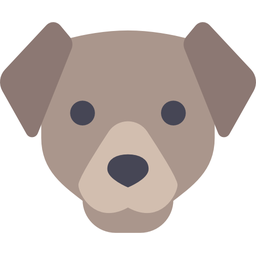} 
    \iconimage{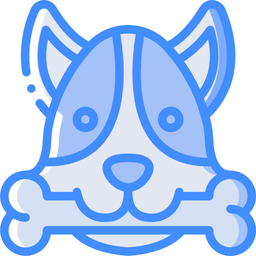}  
    \iconimage{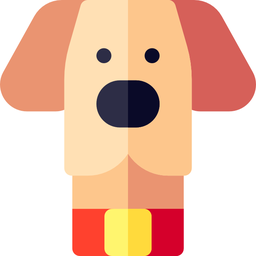}
    \iconimage{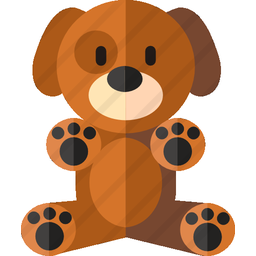}
    \iconimage{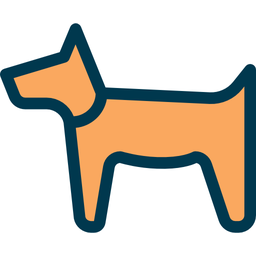} 
    \iconimage{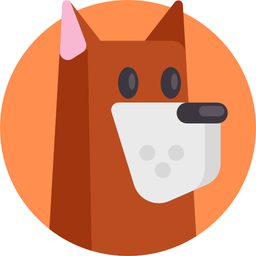} \\ 
    Giraffe & 
    \iconimage{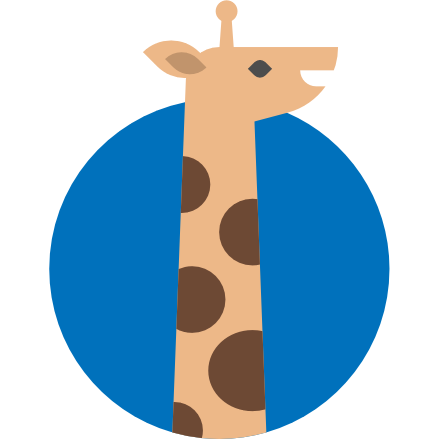} 
    \iconimage{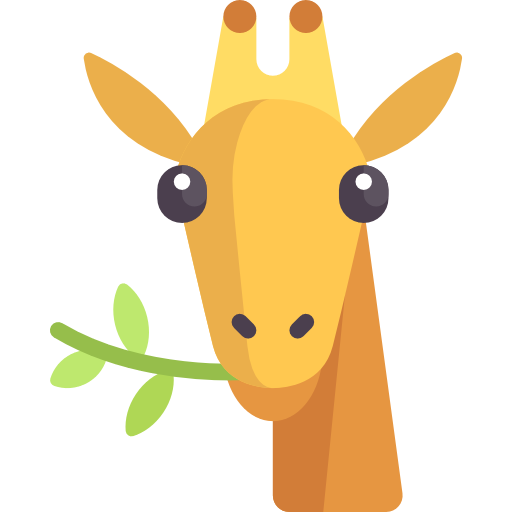} 
    \iconimage{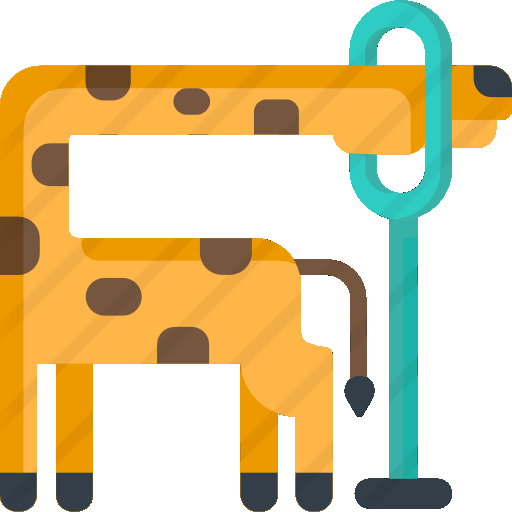} 
    \iconimage{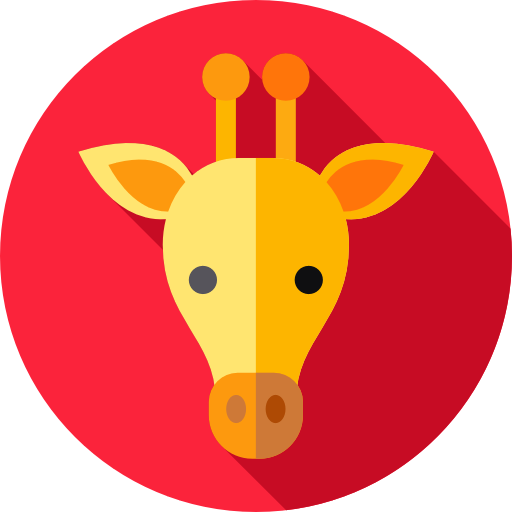}
    \iconimage{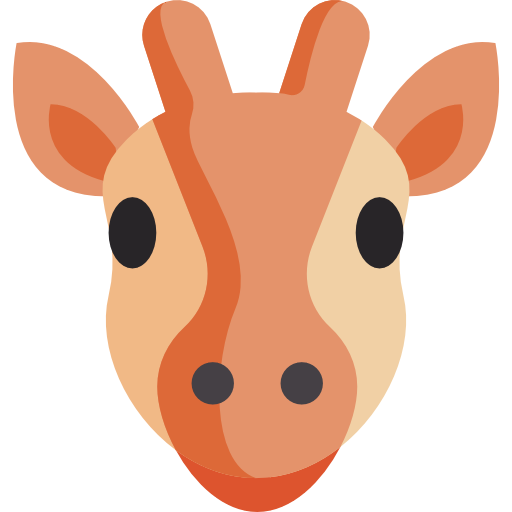}
    \iconimage{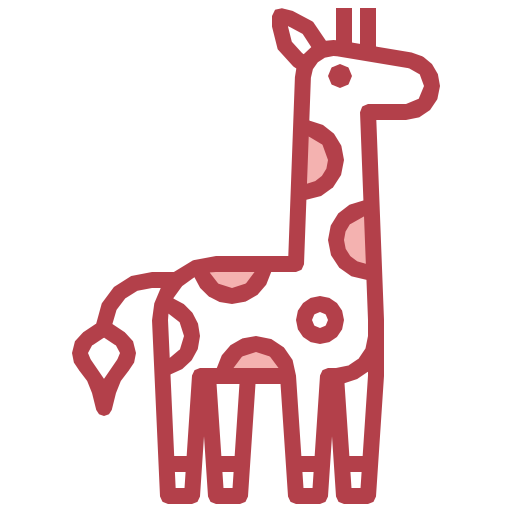}  
    \iconimage{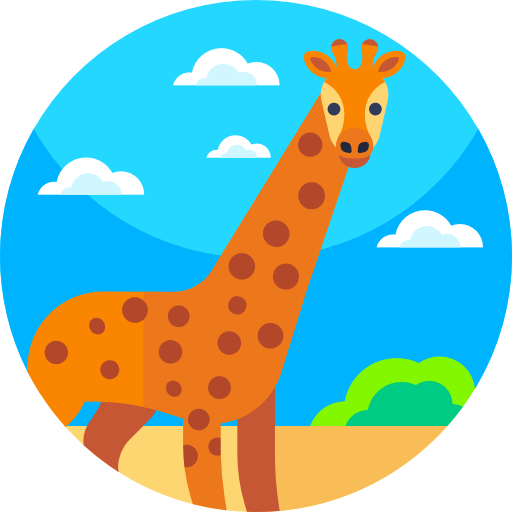} &
    Kite & 
    \iconimage{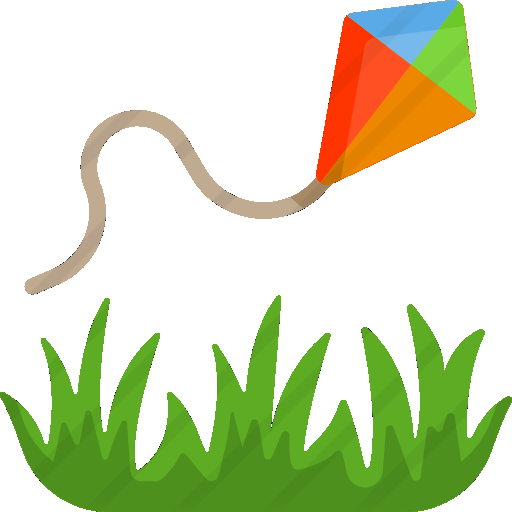} 
    \iconimage{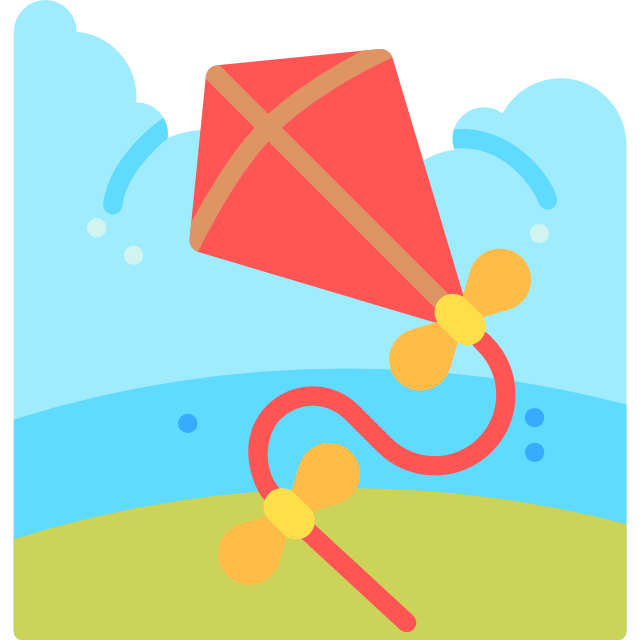} 
    \iconimage{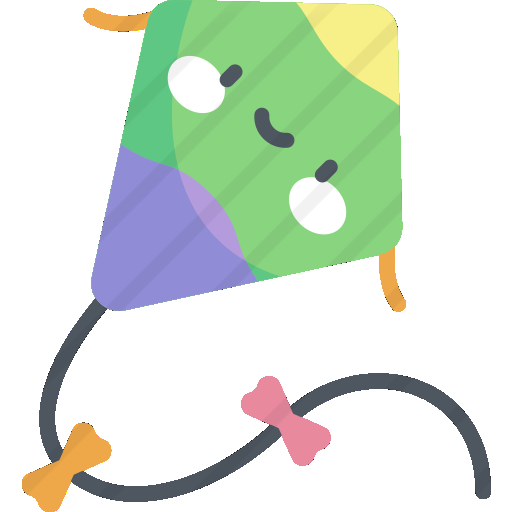} 
    \iconimage{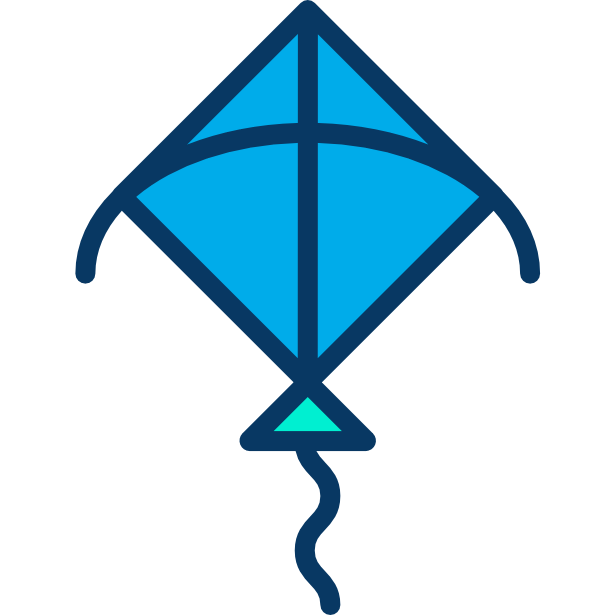}
    \iconimage{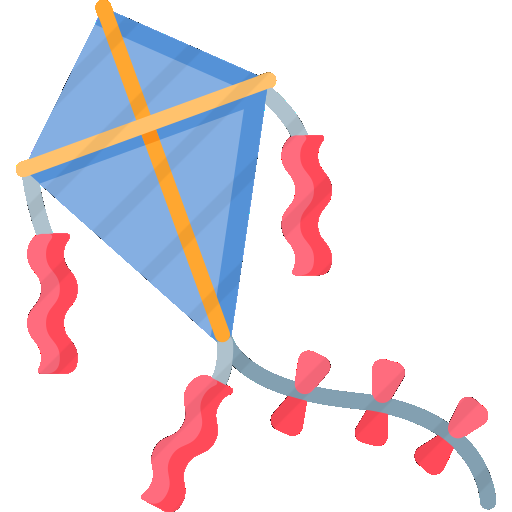}
    \iconimage{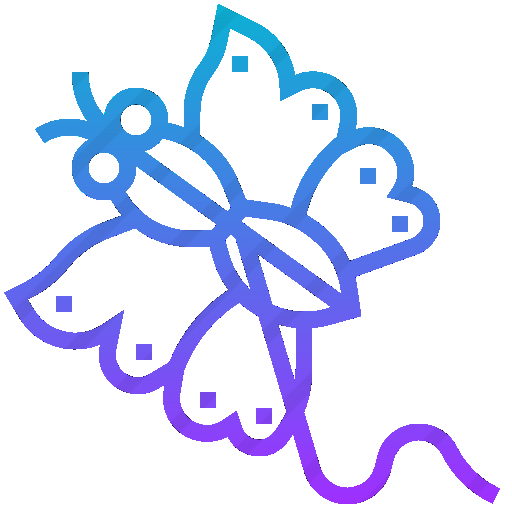}  
    \iconimage{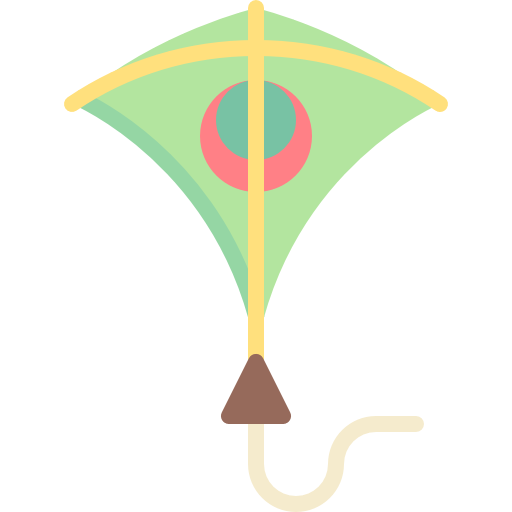}  \\
    Soda & 
    \iconimage{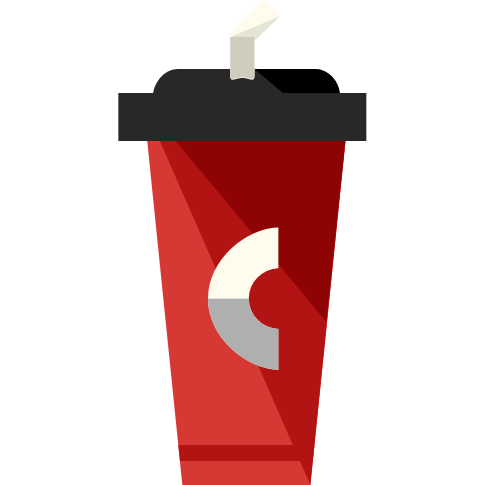} 
    \iconimage{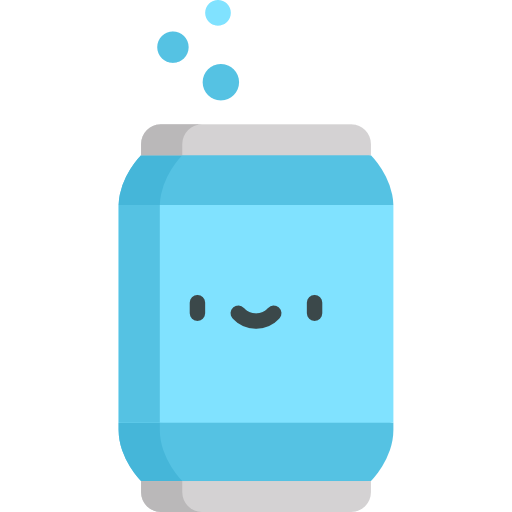} 
    \iconimage{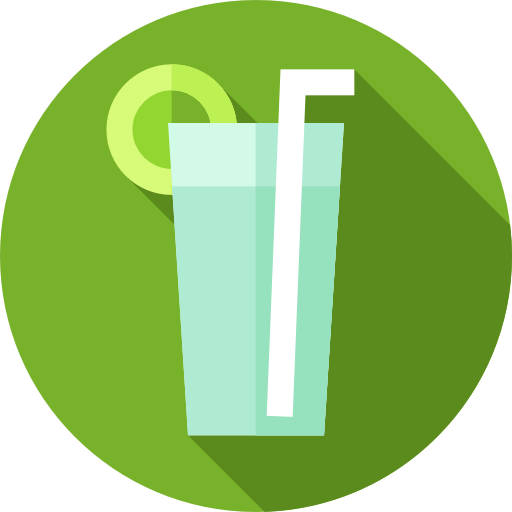} 
    \iconimage{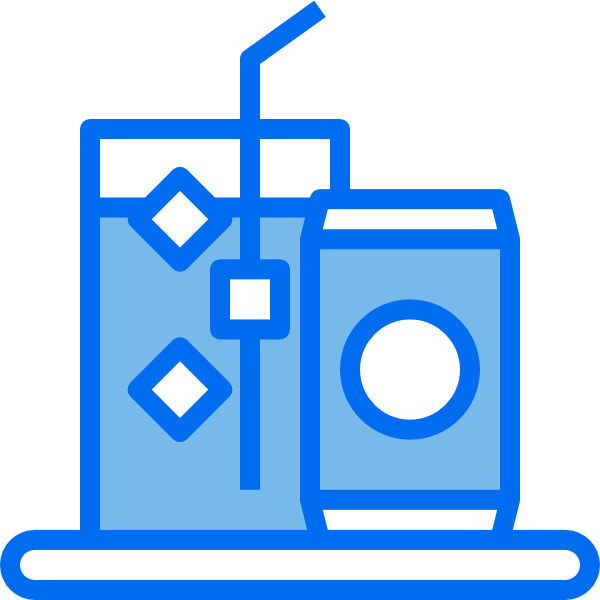}
    \iconimage{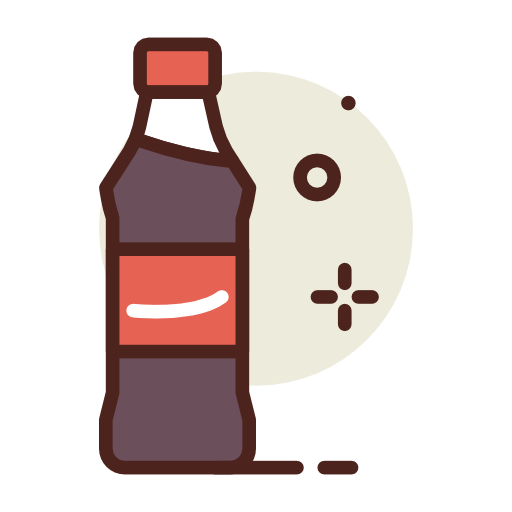}
    \iconimage{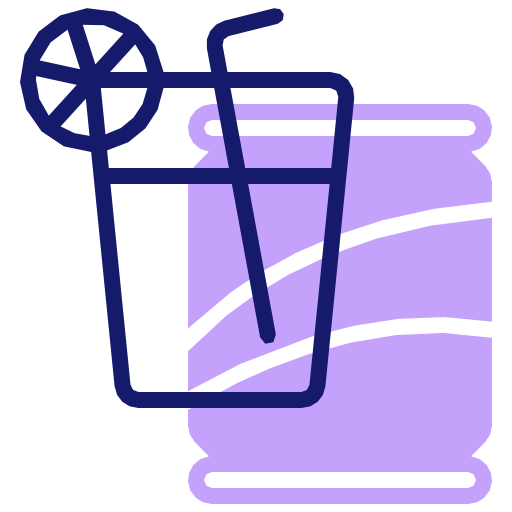}  
    \iconimage{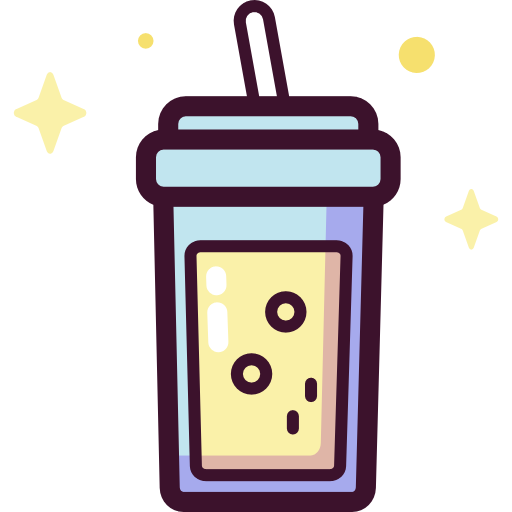}  &
    Tree & 
    \iconimage{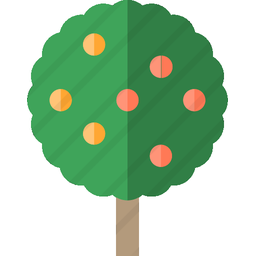}  
    \iconimage{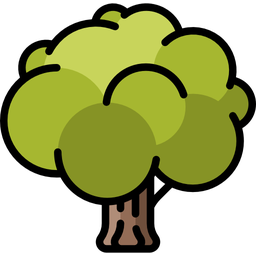} 
    \iconimage{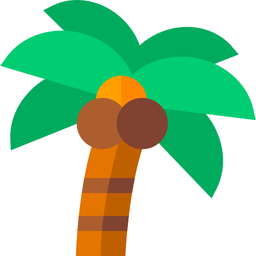}  
    \iconimage{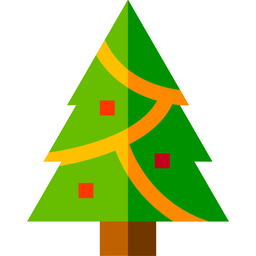}
    \iconimage{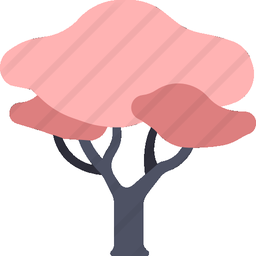}
    \iconimage{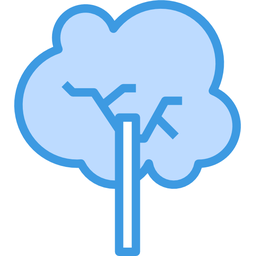}
    \iconimage{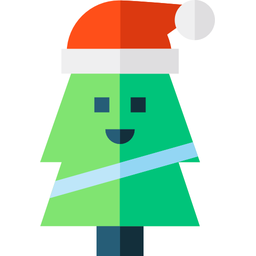}  \\ 
    \bottomrule	
\end{tabular}
\caption{Collected icon examples in the Icon645 dataset.}
\label{table:icon_examples}
\end{table}


\section{Benchmarks}
\label{sec:benchmark}

\begin{figure*}[ht]
    \centering 
    \includegraphics[width= 0.96\linewidth]{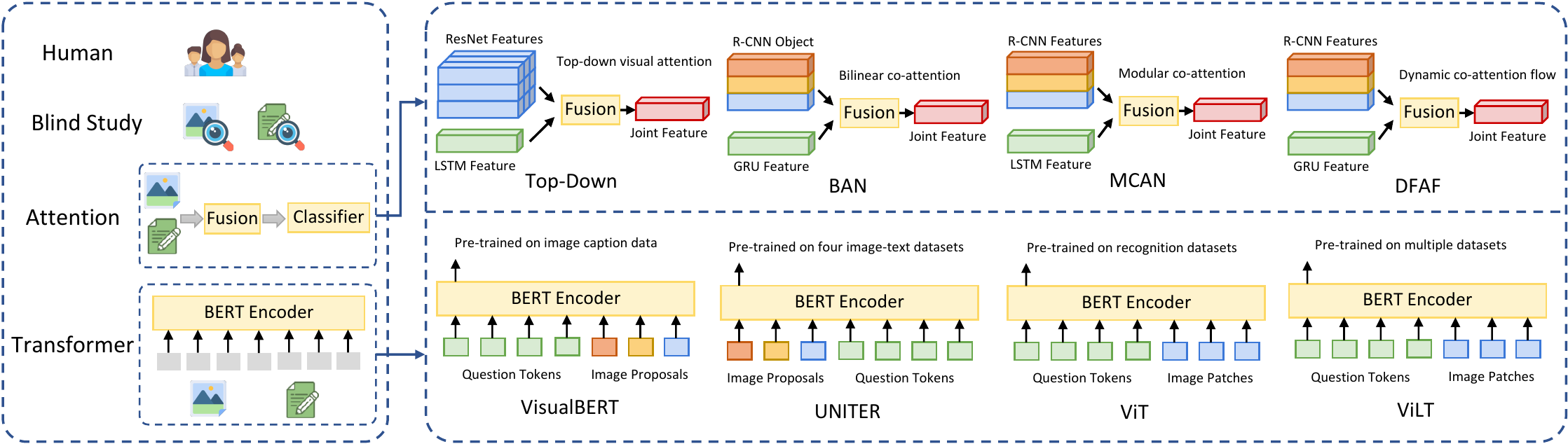}
    \caption{An overview of benchmark baselines on the IconQA task.}
    \label{fig:baselines}
\end{figure*}

In this section, we first develop a patch cross-modal Transformer model (Patch-TRM) as a strong baseline for the IconQA task. To benchmark the IconQA dataset, we consider multi-modal pooling methods with attention mechanisms \cite{Anderson2017up,Kim2018,gao2019dynamic,yu2019mcan}, Transformer-based VQA approaches \cite{lu2019vilbert,chen2020uniter,wonjae2021an,pmlr-v139-kim21k}, and \textcolor{black}{three} blind study methods as benchmark models, as summarized in Figure \ref{fig:baselines}. Additionally, a user study is conducted to explore the performances of human beings in different age groups. In the sections below, we discuss the main principles of the core networks in the benchmarks we performed. 


\subsection{Our Baseline Model}

\begin{figure*}[t]
    \centering 
    \includegraphics[width= 0.92\linewidth]{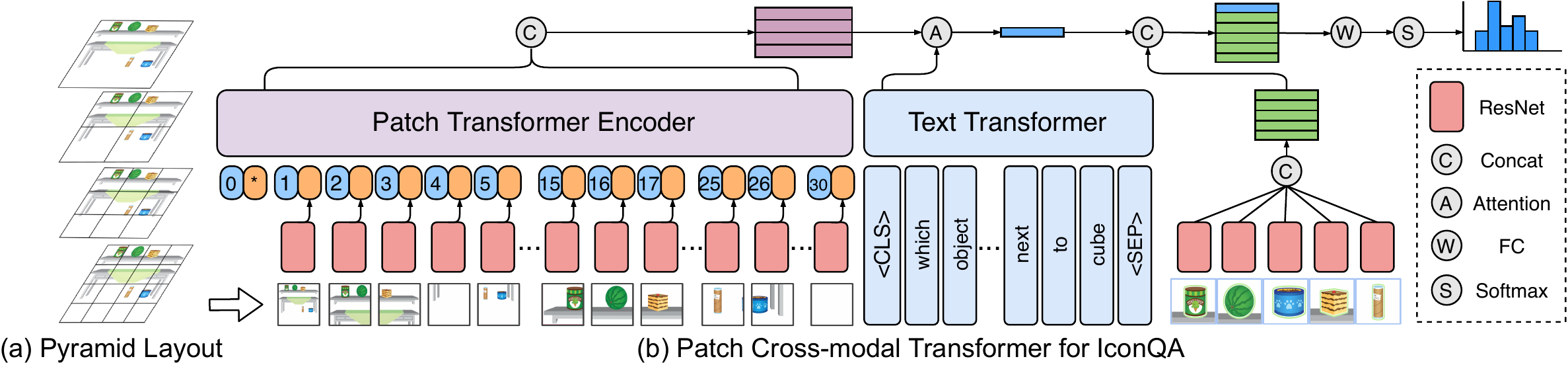}
    \caption{Our IconQA baseline Patch-TRM. Patch-TRM takes patches parsed from a hierarchical pyramid layout and embeds them through ResNet pre-trained on our Icon645 dataset. The joint diagram-question feature is  learned via cross-modal Transformers followed by the attention module.}
    \label{fig:model}
\end{figure*}

Inspired by recent advances Transformer has achieved in vision-language tasks \cite{li2019visualbert,lu2019vilbert}, we develop a cross-modal Transformer model Patch-TRM for icon question answering. Taking the \textit{multi-image choice} sub-task as an example, the overall architecture is shown in Figure \ref{fig:model}. The diagram is first parsed into ordered patches in a hierarchical pyramid layout. These patches are then encoded by a pre-trained ResNet and passed through a vision Transformer. Question text is encoded by a language Transformer and fused with patch embeddings via the attention mechanism. The encoded image choices are concatenated with the joint diagram-question representation and then fed to a classifier for question answering. The other two sub-tasks utilize similar network architectures, except that in the \textit{multi-text-choice} sub-task, we use an LSTM encoder \cite{hochreiter1997long} for choice embedding, while \textit{filling-in-the-blank} does not need a choice encoder. 

Current dominant VQA methods either rely heavily on the ResNet backbone network to extract image features or depend on the Transformer encoders to learn image embeddings. However, these networks are pre-trained on natural images  and are likely to fail to extract meaningful representations or reasonable object proposals when processing the diagrams in IconQA. Instead, we pre-train the ResNet network on the icon classification task with the icon dataset we compiled (Section \ref{sec:Icon645}). Patch-TRM hierarchically parses the diagram into patches that retain complete objects to a large extent, and the parsed patches are embedded by the pre-trained ResNet network before being fed into the vision Transformer. The hierarchical parsing structure, along with the ResNet pre-trained on icon data facilitate our Patch-TRM to learn informative diagram representations for the IconQA task. More details of the pre-training task are discussed in Section \ref{sec:icon_class}.

\subsection{Benchmark Methods}

\textbf{Attention models.} We construct four attention models for benchmarking. The first model implements Top-Down attention \cite{Anderson2017up} for VQA, which is a strong attention method  that applies free-form based attention on image representations from a pre-trained ResNet-101 network. The remaining three models utilize the bottom-up attention mechanism with the help of object detection proposals from Faster-RCNN \cite{ren2015faster}. Specifically, BAN \cite{Kim2018} proposes a method that utilizes bilinear attention distributions to learn joint vision-language information. DFAF \cite{gao2019dynamic} is an advanced model that applies self-attention and cross-modal attention and introduces the information flow to help the model focus on target question words and image regions. The last approach, MCAN \cite{yu2019mcan}, learns the self-attention on the questions and images and the question-guided-attention of images jointly.

\textbf{Transformer models.} Four Transformer-based models are also implemented as benchmarks. ViLBERT \cite{lu2019vilbert} and UNITER \cite{chen2020uniter} are two Transformer-based approaches that take image object proposals from Faster-RCNN \cite{ren2015faster} and question tokens as inputs. Specifically, ViLBERT learns the joint representation of the image content and the natural language content from image proposal regions and question tokens, while UNITER processes multimodal inputs simultaneously for joint visual and textual understanding. The last two benchmarks ViL \cite{wonjae2021an} and ViLT \cite{pmlr-v139-kim21k} are more recently proposed Transformer models that take image patch tokens instead of object proposals as inputs when representing the image.

\textbf{Blind study models.} We develop \textcolor{black}{three} models to check for possible data biases in the IconQA dataset.  \textcolor{black}{A random baseline picks up one from the given choice candidates for the \textit{multiple-choice} sub-tasks while predicts the answer by randomly selecting one from all possible answers in the train data for the \textit{filling-in-the-blank} sub-task.}  Q-Only is set up similar to the Top-Down \cite{Anderson2017up} model, but it only considers textual inputs. This baseline learns the question bias in the training set. I-Only also has a Top-Down architecture, but it only takes abstract diagrams as inputs, and tests the distribution biases in the images and answers in IconQA. 

\textbf{User study.} To assess human performances in the IconQA task, we post the test set of IconQA on Amazon Mechanical Turk (AMT) and ask people to provide answers to the questions in the test set. We also ask the participants to provide us with their age group anonymously. We strongly encourage parents who have young children to let their children complete the questionnaires, as their answers give us insights to how the designed audience of these questions perform. Further details about the user study are included in Appendix \ref{app_user_study}.



\section{Experiments}



\subsection{Training Details}
\label{sec:implementation}

Following prior work \cite{antol2015vqa}, all the baselines are trained on the IconQA training set, tuned on the validation set, and finally evaluated on the test set. Similar to \cite{antol2015vqa}, we choose accuracy as the evaluation metric. For the two \textit{multi-choice} sub-tasks, the answer is regarded as correct only if it matches the ground truth. On the other hand, as the collected answers for \textit{filling-in-blank} are short numbers, correct answers are expanded to include both the digital number and its corresponding words. More details of the benchmark setups and implementations can be found in Appendix \ref{app_experiment_detail}.

Our benchmarks and baselines are implemented using PyTorch. All experiments are run on one Nvidia RTX 3090 GPU. We use the Adamax optimizer with optimal learning rates of $7\times 10^{-4}$, $8 \times 10^{-4}$, and $2 \times 10^{-3}$ on the three sub-tasks respectively. We apply a binary cross-entropy loss to train the multi-class classifier with a batch size of 64 and a maximum epoch of 50. The early stopping strategy is used when the validation accuracy stops improving for five consecutive epochs. It takes about 50, 30, and 10 minutes to train our baseline Patch-TRM on three sub-tasks respectively.

\begin{figure*}[t]
    \centering 
    \includegraphics[width= 0.2068\linewidth]{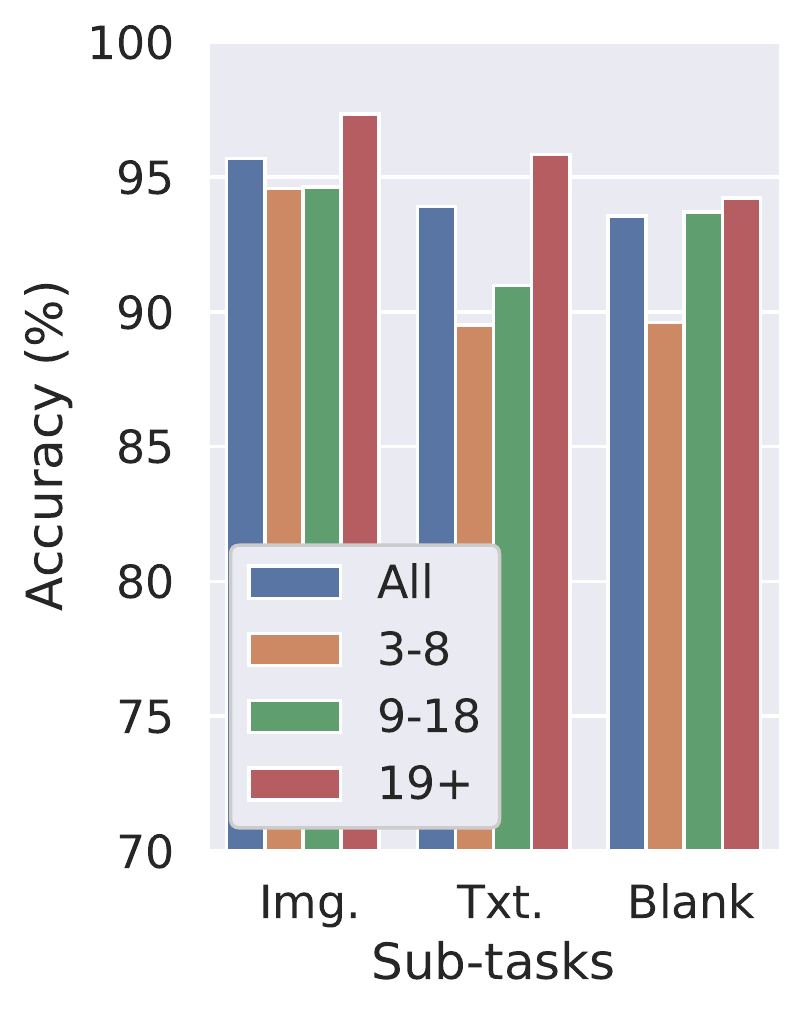}
    \includegraphics[width= 0.69796\linewidth]{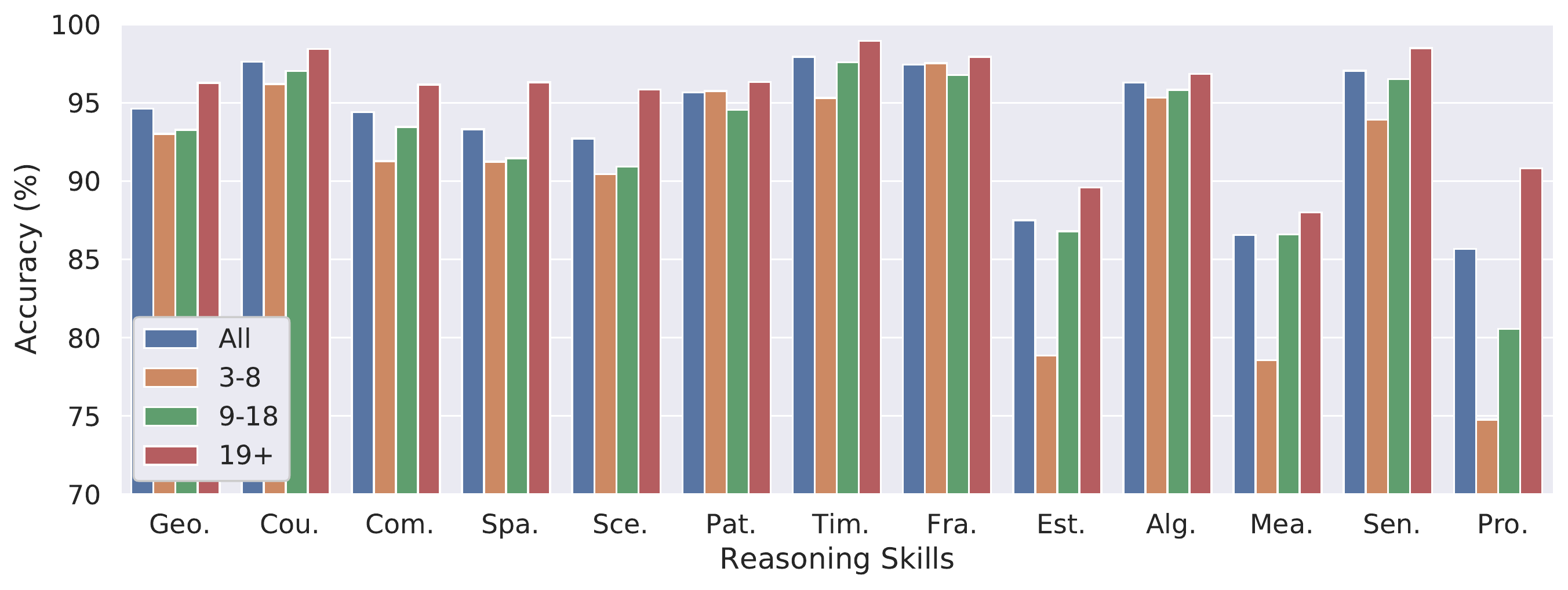}
    \vspace{-1mm}
    \caption{Performance of humans in different age groups for the IconQA task. \textbf{Left}: Accuracy over three sub-tasks; \textbf{Right}: Accuracy over thirteen reasoning skills.}
    \label{fig:humanperf}
\end{figure*}

\subsection{Experimental Results}

\begin{table*}[t]
\centering
\scriptsize
\renewcommand\tabcolsep{2.5pt}
\renewcommand{\arraystretch}{0.91}
\begin{tabular}{{l}*{16}{c}}
    \toprule
    \multirow{3}{*}{} &
    \multicolumn{3}{c}{\textbf{Sub-tasks (3)}} & \multicolumn{13}{c}{\textbf{Reasoning skills (13)}} 	\\
    \cmidrule(lr){2-4} \cmidrule{5-17} 
    \textbf{Method} & Img. & Txt. & Blank 
    & Geo. & Cou. & Com. & Spa. & Sce. & Pat. & Tim. & Fra. & Est. & Alg. & Mea. & Sen. & Pro. \\
    \midrule
    Human & 95.69 & 93.91 & 93.56 & 94.63 & 97.63 & 94.41 & 93.31 & 92.73 & 95.66 & 97.94 & 97.45 & 87.51 & 96.29 & 86.55 & 97.06 & 85.67 \\
    \midrule
    Random & 41.70 &  36.87 &  0.29 &  30.30 & 18.38 & 41.20 &  36.49 &  34.25 &  34.81 &  35.82 &  34.84 &  3.62 & 11.12 &  0.36 &  45.16 &  38.81 \\
    Q-Only & 41.64 & 36.86 & 28.45 & 38.03 & 33.63 & 48.19 & 37.14 & 35.37 & 33.66 & 48.09 & 33.06 & 40.46 & 28.02 & 38.07 & 45.25 & 40.76 \\
    I-Only & 41.56 & 36.02 & 46.65 & 38.71 & 37.64 & 45.26 & 37.52 & 35.47 & 36.29 & 47.37 & 32.48 & 62.29 & 31.73 & 64.02 & 45.25 & 37.51 \\
    \midrule
    Top-Down \cite{Anderson2017up} & 75.92 & 68.51 & 73.03 & 80.07 & 65.01 & 80.65 & 45.78 & 58.22 & 55.01 & 68.28 & 72.43 & \textbf{99.54} & 50.00 & \textbf{99.46} & 84.54 & 83.75 \\
    BAN \cite{Kim2018} & 76.33 & 70.82 & 75.54 & 79.99 & 67.56 & 82.12 & 53.20 & 66.92 & 55.67 & 66.50 & 73.77 & 97.06 & 47.46 & 96.50 & 82.12 & 82.45 \\
    ViLBERT \cite{li2019visualbert} & 76.66 & 70.47 & 77.08 & 80.05 & 71.05 & 75.60 & 49.46 & 58.52 & 62.78 & 66.72 & 74.09 & 99.22 & 50.62 & 99.07 & 81.78 & 70.94 \\
    MCAN \cite{yu2019mcan} & 77.36 & 71.25 & 74.52 & 79.86 & 68.94 & 82.73 & 49.70 & 62.49 & 54.79 & 68.00 & 76.20 & 99.08 & 47.32 & 98.99 & 83.25 & 84.87 \\
    DFAF \cite{gao2019dynamic} & 77.72 & 72.17 & 78.28 & 81.80 & 70.68 & 81.69 & 51.42 & \textbf{67.01} & 56.60 & 67.72 & 77.60 & 99.02 & 50.27 & 98.83 & 84.11 & 85.70 \\
    UNITER \cite{chen2020uniter} & 78.71 & 72.39 & 78.53 & 81.31 & 71.01 & 83.67 & 48.34 & 61.25 & 60.81 & 69.77 & 78.37 & 99.41 & 49.18 & 99.38 & 86.10 & 87.84 \\
    ViT \cite{wonjae2021an} & 79.15 & 72.34 & 78.92 & 82.60 & 70.84 & 82.12 & 54.64 & 68.80 & 58.46 & 68.66 & 77.41 & 98.95 & 51.10 & 98.76 & 84.72 & 86.07 \\
    ViLT \cite{pmlr-v139-kim21k}  & 79.67 & 72.69 & 79.27 & \textbf{82.61} & 71.13 & 84.95 & 53.38 & 66.72 & 59.22 & 69.99 & 75.81 & 99.02 & 50.55 & 98.91 & 86.10 & 87.65 \\
    \midrule
    \textbf{Patch-TRM} (Ours) & \textbf{82.66} & \textbf{75.19} & \textbf{83.62} & 81.87 & \textbf{77.81} & \textbf{87.00} & \textbf{55.62} & 62.39 & \textbf{68.75} & \textbf{77.98} & \textbf{82.13} & 98.24 & \textbf{56.73} & 97.98 & \textbf{92.49} & \textbf{95.73} \\
    \bottomrule	
\end{tabular}
\caption{Results on the IconQA dataset.}
\label{table:result}
\end{table*}

Table \ref{table:result} demonstrates the results of the benchmark methods and our baseline on the IconQA test set. The first three columns of the results represent the three sub-tasks: \textit{multi-image-choice}, \textit{multi-text-choice}, and \textit{filling-in-the-blank} respectively. The remaining 13 columns illustrate the results of these approaches over problems that require different reasoning skills, as defined in Table \ref{table:skills}.

\textbf{Human performance.} Out of the 54,896 collected answers, 9,620 are made by young children from age 3 to 8, 19,040 are made by adolescents from age 9 to 18, and 26,236 are made by adults. The human performance over the three sub-tasks and thirteen skills is illustrated in Figure \ref{fig:humanperf}. As expected, young children do not answer the questions as well as adolescents or adults, suggesting that most participants answer their ages correctly. Moreover, the result shows that humans perform more consistently on all sub-tasks compared to machine algorithms. Interestingly, humans are outperformed by models quite significantly in questions that require numerical reasoning skills like \textit{probability}, \textit{measurement}, and \textit{estimation}.


\textbf{Analysis by Task Types.} Humans outperform all benchmarks consistently over there sub-tasks and most reasoning skills. There is still a large gap to fill for future research of abstract diagram understanding and visual reasoning on the icon domain. The results achieved in blind studies of Q-only and I-only are close to random, showing that the IconQA dataset is robust and reliable in distribution. Our proposed Patch-TRM baseline outperforms current state-of-the-art VQA models in all three sub-tasks. These improvements mainly come from two insights: pre-training ResNet on icon images and taking a hierarchical approach with attention mechanism.




\textbf{Analysis by Reasoning Types.} Similarly, the Patch-TRM baseline obtains better results than the benchmarks over most reasoning skill types. Interestingly, in some skills such as \textit{estimation}, \textit{measurement}, and \textit{probability}, Patch-TRM performs better than average human beings. This implies that neural networks have a promising potential to develop the basic ability of mathematical reasoning.

\textbf{Quantitative Analysis.} We visualize one example with the cross-modal attention map generated by our baseline Patch-TRM in Figure \ref{fig:attention}. The visualized attention shows that our baseline is capable of attending to the corresponding patch regions with higher weights given the input question. 




\subsection{Ablation Study}
\begin{figure}[t]
    \centering
    \scriptsize
    \renewcommand\tabcolsep{9.0pt}
    \begin{minipage}{0.52\textwidth}     
    \centering 
        \includegraphics[width=0.89\textwidth]{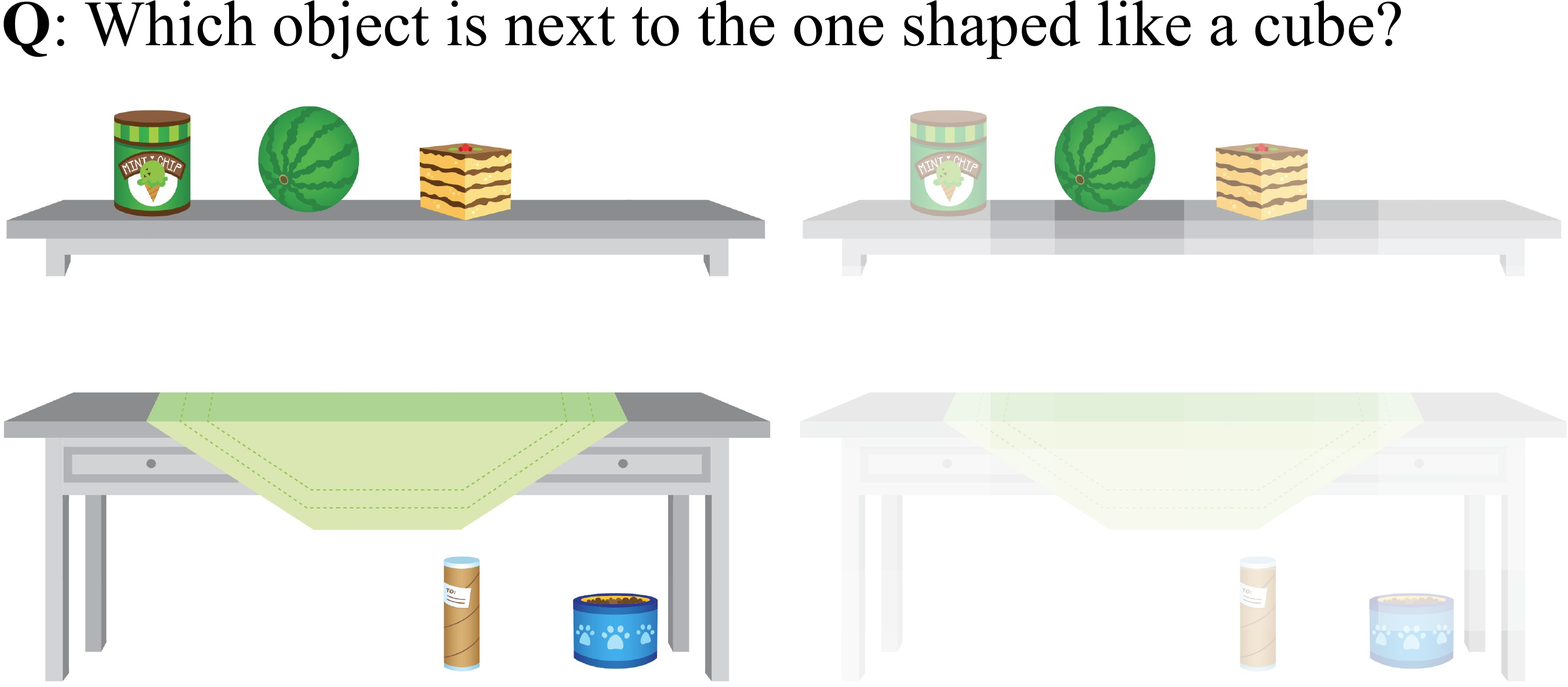} 
        \caption{Text-to-image attention visualization.}
        \label{fig:attention} 
    \end{minipage}
    \hfill
    \begin{minipage}{0.46\textwidth} 
    \begin{tabular}{{l}*{3}{c}}
        \toprule
        \textbf{Method} & \text{Img.} & \text{Txt.} & \text{Blank} \\
        \midrule
        Patch-TRM w/o pre & 82.01 & 72.72 & 81.67 \\  
        Patch-TRM w/o att & 80.63 & 68.00 & 80.29 \\ 
        Patch-TRM w/o pos & 81.27 & 64.98 & 80.68 \\  
        Patch-TRM V-CLS   & 80.15 & 63.90 & 70.27 \\  
        Pyramid 1+4+9+16 & 82.45 & 68.76 & 82.19 \\  
        Pyramid 1+4+9    & 80.61 & 67.42 & 81.36 \\  
        \midrule
        \textbf{Full model} & \textbf{82.96} & \textbf{75.21} & \textbf{83.10}  \\
        \bottomrule	
    \end{tabular}
    \captionof{table}{Ablation study in IconQA.}
    \label{table:ablation}
    \end{minipage} 
\end{figure}

To study the functions of individual components in our model, we conduct an ablation analysis. Table \ref{table:ablation} presents the results of different simplifications of our full model, where each implementation is trained on the IconQA train set and tested on the validation set. Instead of ResNet101 pre-trained on the icon classification task, \textit{Patch-TRM w/o pre} utilizes ResNet101 pre-trained on natural image data for patch feature extraction. The decreasing performance of 0.95-2.49\% indicates that pre-training backbones on tasks within similar domains is critical to downstream tasks. The attention mechanism helps to combine the image and question representations and improves the model performance by up to 7\% compared to using simple concatenation (denoted as \textit{Patch-TRM w/o att}). Positional embeddings of the ordered diagram patches benefit the vision Transformer by enabling it to learn spatial relationships among the patches, compared to the baseline without position embeddings (\textit{Patch-TRM w/o pos}). \textit{Patch-TRM V-CLS} uses the output embedding of $\mathtt{[CLS]}$ token as the diagram feature instead, which leads to a drastic performance decline. We have also experimented with coarse-grained patch cropping (e.g., \textit{Pyramid 1+4+9+16} denotes 30 patches, \textit{Pyramid 1+4+9} denotes 14 patches), which results in  a performance degradation of 0.51\% to 7.79\%.

\subsection{Icon Classification for Pre-training}
\label{sec:icon_class}

The Icon645 dataset is collected to pre-train the backbone network for patch feature extraction. 
\begin{wraptable}{r}{6.2cm}
    \centering
    \scriptsize
    \renewcommand\tabcolsep{1.0pt}
    \vspace{1mm}
    \begin{tabular}{{l}*{4}{c}}
        \toprule
        \textbf{Method} & Total & Head & Medium & Tail \\
        \midrule
        ResNet32 \cite{he2016deep}  + CB \cite{cui2019class} & 27.91 & 19.66 & 36.51 & 33.53 \\
        ResNet32 \cite{he2016deep} + Focal Loss \cite{lin2017focal}  & 32.80 & 51.59 & 36.51 & 8.94 \\ 
        ResNet32 \cite{he2016deep} + LDAM \cite{cao2019learning}      & 42.65 & 55.68 & 46.42 & 24.94 \\
        ResNet101 \cite{he2016deep} + LDAM \cite{cao2019learning}   & \textbf{62.93} & \textbf{70.29} & \textbf{70.50} & \textbf{47.51} \\ 
        \bottomrule	
    \end{tabular}
    \caption{Results for icon classification.}
    \label{table:icon_classifier}
\end{wraptable}
The dataset has a long-tailed distribution, and thus we address the class-imbalanced issue following previous studies on specific loss functions such as CB loss \cite{cui2019class}, Focal loss\cite{lin2017focal}, and LDAM loss \cite{cao2019learning}. 
The metric of Top-5 accuracy is used to evaluate different model setups and the evaluation results are summarized in Table \ref{table:icon_classifier}. 
Following \cite{liu2019large}, to demonstrate performances on different data parts, we divide the dataset into three balanced clusters: Head, Medium, and Tail, corresponding to 132, 122, and 123 classes respectively. \textcolor{black}{All classes in Head have at least 1,000 instances, all classes in Medium have 300 - 999 instances, and all classes in Tail have fewer than 300 instances.}  As the results show, the backbone network ResNet101 with a re-balanced LDAM loss function achieves the best result for icon classification on Icon645. Consequently, we adopt this pre-trained ResNet101 network to extract patch features in our baseline Patch-TRM for IconQA.




\section{Conclusion}

In this work, we introduce IconQA, an open-source dataset of icon question answering in real-world scenarios for assessing the abilities of abstract diagram understanding and visual language reasoning. IconQA features 107,439 questions, three sub-tasks, and thirteen types of cognitive reasoning skills. We benchmark the IconQA task extensively with a user study, \textcolor{black}{three} blind studies, as well as multiple existing attention-based and Transformer-based approaches. We further develop a strong baseline, Patch-TRM, which parses the diagram in a pyramid layout and applies cross-modal Transformers with attention mechanism to learn the meaningful joint diagram-question feature. Additionally, we introduce Icon645, a large-scale icon dataset that is useful to pre-train the diagram encoding network used in Patch-TRM for the IconQA task. 

By releasing a new dataset of icon question answering for abstract diagram understanding and visual language reasoning, we envision that IconQA will facilitate a wide range of research in computer vision and natural language processing, as well as smart education applications like tutoring systems, to invent the future of AI for science education.


{
\small
\bibliography{egbib}
}

\newpage
\section*{Checklist}


\begin{enumerate}

\item For all authors...
\begin{enumerate}
  \item Do the main claims made in the abstract and introduction accurately reflect the paper's contributions and scope?
    \answerYes{}
  \item Did you describe the limitations of your work?
    \answerYes{} Please see Appendix \ref{app:limitation}.
  \item Did you discuss any potential negative societal impacts of your work?
    \answerYes{} Please see Section \ref{sec:impact}.
  \item Have you read the ethics review guidelines and ensured that your paper conforms to them?
    \answerYes{}
\end{enumerate}

\item If you are including theoretical results...
\begin{enumerate}
  \item Did you state the full set of assumptions of all theoretical results?
    \answerNA{}
	\item Did you include complete proofs of all theoretical results?
    \answerNA{}
\end{enumerate}

\item If you ran experiments (e.g. for benchmarks)...
\begin{enumerate}
  \item Did you include the code, data, and instructions needed to reproduce the main experimental results (either in the supplemental material or as a URL)?
    \answerYes{} Please see our project page at \url{https://iconqa.github.io}.
  \item Did you specify all the training details (e.g., data splits, hyperparameters, how they were chosen)?
    \answerYes{} Please see Section \ref{sec:implementation} for training details. For details on benchmark model settings, see Appendix \ref{app_experiment_detail}.
	\item Did you report error bars (e.g., with respect to the random seed after running experiments multiple times)?
    \answerNA{}
	\item Did you include the total amount of compute and the type of resources used (e.g., type of GPUs, internal cluster, or cloud provider)?
    \answerYes{} Please see Section \ref{sec:implementation}. 
\end{enumerate}

\item If you are using existing assets (e.g., code, data, models) or curating/releasing new assets...
\begin{enumerate}
  \item If your work uses existing assets, did you cite the creators?
    \answerYes{}
  \item Did you mention the license of the assets?
    \answerYes{} See \url{https://github.com/lupantech/IconQA#license}.
  \item Did you include any new assets either in the supplemental material or as a URL?
    \answerYes{} All datatsets are available on the IconQA website \url{https://iconqa.github.io}, or the github repository \url{https://github.com/lupantech/IconQA}.
  \item Did you discuss whether and how consent was obtained from people whose data you're using/curating?
    \answerNA{}
  \item Did you discuss whether the data you are using/curating contains personally identifiable information or offensive content?
    \answerYes{} As we discuss in Section \ref{sec:impact}, our datasets do not contain identifiable or offensive content.
\end{enumerate}

\item If you used crowdsourcing or conducted research with human subjects...
\begin{enumerate}
  \item Did you include the full text of instructions given to participants and screenshots, if applicable?
    \answerYes{} See Appendix \ref{app_user_study}.
  \item Did you describe any potential participant risks, with links to Institutional Review Board (IRB) approvals, if applicable?
    \answerNA{}
  \item Did you include the estimated hourly wage paid to participants and the total amount spent on participant compensation?
    \answerYes{} See Appendix \ref{app:compensation}.
\end{enumerate}

\end{enumerate}


\newpage
\appendix

\hrule height 4pt
\vskip 0.25in
\vskip -\parskip
\vbox{
    \centering
    \LARGE 
    \textbf{Supplementary Materials for \\  
    IconQA: A New Benchmark for Abstract Diagram Understanding and Visual Language Reasoning}
}
\vskip 0.29in
\vskip -\parskip
\hrule height 1pt
\vskip 0.09in
\vskip 0.5in
\renewcommand*\footnoterule{} 

\newcommand\blfootnote[1]{%
  \begingroup
  \renewcommand\thefootnote{}\footnote{#1}%
  \addtocounter{footnote}{-1}%
  \endgroup
}

\blfootnote{
    \footnotesize 
    \\35th Conference on Neural Information Processing Systems (NeurIPS 2021) Track on Datasets and Benchmarks.
}

\section{The IconQA Dataset}
The following datasheet follows the format suggested in this paper \cite{gebru2018datasheets}.

\label{app_iconqa_data}

\subsection{More Examples}
\label{app_more_examples}

\begin{figure*}[ht!]
\centering 
    \includegraphics[width= 0.98\linewidth]{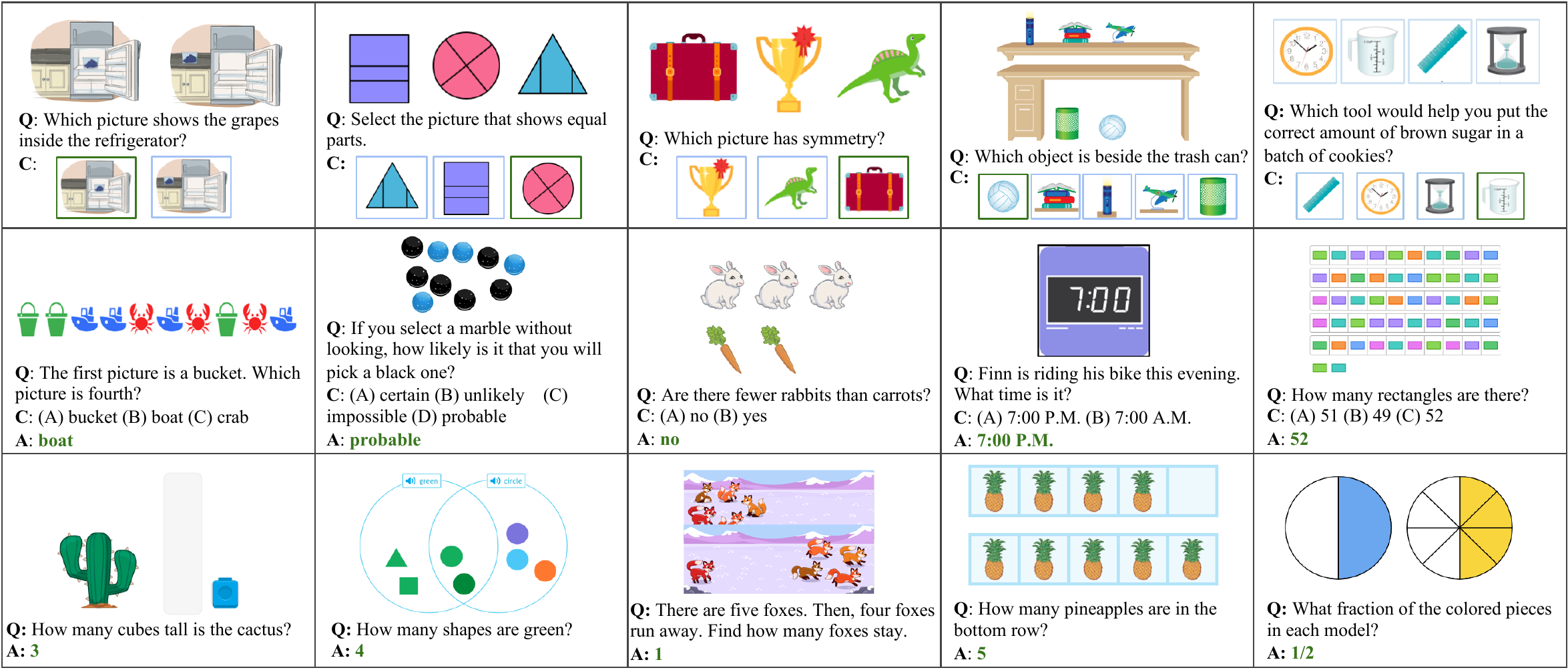}
    \vspace{0mm}
    \caption{More examples in the IconQA dataset. \textbf{Top}: The \textit{multi-image-choice} sub-task. \textbf{Middle}: The \textit{multi-text-choice} sub-task. \textbf{Bottom}: The \textit{filling-in-the-blank} sub-task.}
    \vspace{-1mm}
\label{fig2:more_examples}
\end{figure*}

\subsection{Dataset Label}
The IconQA dataset label is shown in Figure \ref{fig:card_iconqa}.
\begin{figure}[ht]
    \centering
    \resizebox{0.6\linewidth}{!}{
        \sffamily
{
\fbox{%
    \begin{tabular}{@{}p{\NFwidth}@{}}
    
    \multicolumn{1}{c}{\huge\bfseries IconQA Dataset Facts}\\
    \NFrule
    \NFelement{Website}{}{ \url{https://iconqa.github.io}}\\
    
    \NFRULE
    \NFline{Metadata}\\
    \NFrule
    \NFelement{Instances}{}{107,439}\\
    \NFelement{Format}{}{.png, .json}\\
    \NFelement{License}{}{CC BY-NC-SA}\\
    \NFelement{Original Use Case}{}{Training VQA systems}\\
    
    \NFRule
    \NFline{Composition}\\
    \NFrule
    \NFelement{Sample or Complete}{}{Sample from \text{ixl.com}}\\
    \NFelement{Missing Data}{}{No data is missing}\\
    
    \NFRule
    \NFline{Collection}\\
    \NFrule
    \NFelement{Sampling Strategy}{}{See the main paper}\\
    \NFelement{Author Consent}{}{None needed}\\
    
    \NFRule
    \NFline{Cleaning and Labeling}\\
    \NFrule
    \NFelement{Cleaning Done}{}{Repetitions and redundancies removed}\\
    \NFelement{Labeling Done}{}{Yes}\\
    
    \NFRule
    \NFline{Uses and Distribution}\\
    \NFrule
    \NFelement{Notable Uses}{}{Training VQA systems}\\
    \NFelement{Original Distribution}{}{Check dataset website}\\
    
    \NFRule
    \NFline{Maintenance and Evolution}\\
    \NFrule
    \NFelement{Corrections or Erratum}{}{None as of now}\\
    \NFelement{Methods to Extend}{}{Contact the author}\\
    
    \NFRULE
    \NFline{Breakdown \hfill\% of Example}\\
    \NFrule
    \NFelement{multi-image-choice}{57,672\,items}{53.7\%}\\
    \NFrule
    \NFelement{multi-text-choice}{31,578\,items}{29.4\%}\\
    \NFrule
    \NFelement{fill-in-the-blank}{18,189\,items}{16.9\%}\\
    \NFrule
    \end{tabular}}
\par} 
        }
    \caption{IconQA dataset label, created with the template from the paper \cite{bandy2021addressing}.}
    \label{fig:card_iconqa}
    \vspace{-1mm}
\end{figure}

\subsection{Question Skill Categories}
The questions we collected contain meta-information including question topics, chapter names, image names, etc. After extensive data exploration by well-informed individuals, we designed a set of rules that map each question to 1-3 of the 13 categories based on trigger words in metadata. The rules for trigger words are list in Table \ref{table:skill_rules}.

\begin{table}[ht!]
    \centering
    \small
    \caption{Trigger words in metadata for skill categories.}
    \begin{tabularx}{\textwidth}{lX}
        \toprule
        \textbf{Skill types} & \text{Trigger words in metadata} \\
        \midrule
        Geometry & name the shape, shapes of, classify shapes, solid, corners, faces, edges, vertices, sides, dimensional, rectangle, circle, triangle, square, rhombus, sphere, cylinder, cone,  cubes, hexagon, perimeter, area, curved, open and close, flip turn, symmetry \\
        Counting & count, tally, a group, ordinal number, area, even or odd, place value, represent numbers, comparing review, equal sides, square corners, one more, one less, fewer, enough, more. \\
        Comparing & compare, comparing, more, less, fewer, enough, wide and narrow, light and heavy, long and short, tall and short, match analog and digital \\
        Spatial & top, above, below, beside, next to, inside and outside, left \\
        Scene & problems with pictures, beside, above, inside and outside, wide and narrow, objects \\
        Pattern & the next, comes next, ordinal number, different \\
        Time & clock, am or pm, elapsed time, times \\
        Fraction & equal parts, halves, thirds, fourths, fraction \\
        Estimation & estimate, measure \\
        Algebra & count to fill, skip count, tally, even or odd, tens and ones, thousands, of ten, elapsed time, perimeter, area, divide \\
        Measurement & measure \\
        Commonsense & light and heavy, compare size, holds more or less, am or pm, times of, tool \\
    Probability & likely \\ \bottomrule
    \end{tabularx}
    \vspace{-3mm}
    \label{table:skill_rules}
\end{table}
\color{black}

\subsection{Links}

The link to download the IconQA dataset can be found at \url{iconqa.github.io}.

\subsection{Motivation}

\begin{itemize}
    \item \ques{For what purpose was the dataset created? Was there a specific task in mind? Was there a specific gap that needed to be filled?}
    \begin{itemize}
        \item IconQA is created to provide researchers with a wide range of VQA data on the abstract image domain. Currently, existing datasets 1) are limited to natural images, or 2) contain diagrams generated with templates, and therefore lack linguistic variation, or 3) include too much domain specific knowledge. We believe that no other abstract diagram QA dataset exists that covers such a wide range of perceptive and cognitive abilities without requiring complicated domain knowledge.
    \end{itemize}
    \item \ques{Who created the dataset (e.g., which team, research group) and on
behalf of which entity (e.g., company, institution, organization)?}
    \begin{itemize}
        \item This dataset was created under the combined effort of multiple researchers from University of California, Los Angeles, Sun Yat-sen University, East China Normal University, and Columbia University.
    \end{itemize}
    \item \ques{Who funded the creation of the dataset?}
    \begin{itemize}
        \item The project received no funding or associated grant.
    \end{itemize}
\end{itemize}

\subsection{Composition}

\begin{itemize}
    \item \ques{What do the instances that comprise the dataset represent (e.g.,
documents, photos, people, countries)? Are there multiple types of
instances (e.g., movies, users, and ratings; people and interactions between them; nodes and edges)?}
    \begin{itemize}
        \item Each instance is a complete icon question answering task.
    \end{itemize}
    \item \ques{How many instances are there in total (of each type, if appropriate)?}
    \begin{itemize}
        \item There are a total of 107,439 instances. 57,672 are \textit{multi-image-choice} questions, 31,578 are \textit{multi-text-choice} questions, and 18,189 are \textit{filling-in-the-blank} questions.
    \end{itemize}
    \item \ques{Does the dataset contain all possible instances or is it a sample
(not necessarily random) of instances from a larger set?}
    \begin{itemize}
        \item The dataset does not contain all possible instances.
    \end{itemize}
    \item \ques{What data does each instance consist of? }
    \begin{itemize}
        \item Each instance in IconQA includes a textual question, an image, and multiple optional visual / textual choices. We also included some metadata about each question, such as the skill type, question type, etc.
    \end{itemize}
    \item \ques{Is there a label or target associated with each instance?}
    \begin{itemize}
        \item Yes, each question is associated with a ground truth answer.
    \end{itemize}
    \item \ques{Is any information missing from individual instances?}
    \begin{itemize}
        \item No. All related information is included in the dataset.
    \end{itemize}
    \item \ques{Are there recommended data splits (e.g., training, development/validation, testing)?}
    \begin{itemize}
        \item Yes. Following conventions in the field, we have splitted the dataset into a training set, a validation set, and a test set with a 0.6:0.2:0.2 ratio.
    \end{itemize}
    \item \ques{Are there any errors, sources of noise, or redundancies in the
dataset?}
    \begin{itemize}
        \item We randomly selected 1,000 questions from each sub-task and ask an experienced expert to double check the answers carefully. Among the 1,000 \textit{multi-image-choice} questions, only 1 error was found. Among the 2,000 questions of the other two sub-tasks, no error was found.
        \item In the \textit{multi-image-choice} sub-task, questions that ask ``Which two are exactly the same?'' might be a source of noise for certain use cases, as in the data label, only one correct answer out of the two is given. We intend to address this problem in the later versions of the dataset.
    \end{itemize}
    \item \ques{Is the dataset self-contained, or does it link to or otherwise rely on
external resources (e.g., websites, tweets, other datasets)?}
    \begin{itemize}
        \item The dataset is self-contained. All related information is included in the dataset.
    \end{itemize}
    \item \ques{Does the dataset contain data that might be considered confidential?}
    \begin{itemize}
        \item No, the dataset does not contain anything related to any individual.
    \end{itemize}
    \item Does the dataset contain data that, if viewed directly, might be offensive, insulting, threatening, or might otherwise cause anxiety?
    \begin{itemize}
        \item No, the dataset does not contain anything offensive.
    \end{itemize}
\end{itemize}

\subsection{Collection Process}
\begin{itemize}
    \item How was the data associated with each instance acquired?
    \begin{itemize}
        \item The data is publicly available on ixl.com. More details are included in the main paper.
    \end{itemize}
    \item What mechanisms or procedures were used to collect the data (e.g., hardware apparatus or sensor, manual human curation, software program, software API)?
    \begin{itemize}
        \item We implemented an integrated graphic user interface tool using Python to help crowd workers to collect the data.
    \end{itemize}
    \item Over what timeframe was the data collected?
    \begin{itemize}
        \item The dataset was finally completed in March, 2021 after three months of data collection, cleaning, and prepossessing.
    \end{itemize}
    \item Were any ethical review processes conducted (e.g., by an institutional review board)?
    \begin{itemize}
        \item No, we did not conduct an ethical review under the assumption that math and science questions designed for young children do not contain any discriminative or offensive content.
    \end{itemize}
    \item Does the dataset relate to people?
    \begin{itemize}
        \item No, the dataset does not relate to people.
    \end{itemize}
\end{itemize}

\subsection{Preprocessing}

\begin{itemize}
    \item Was any preprocessing/cleaning/labeling of the data done (e.g., discretization or bucketing, tokenization, part-of-speech tagging, SIFT feature extraction, removal of instances, processing of missing values)?
    \begin{itemize}
        \item We cropped white space from each diagram in IconQA to tighten it up. Questions with invalid diagrams, answers, or choices were filtered out. Redundant instances were removed based on the metrics of exact question text matching and diagram similarity.
    \end{itemize}
    \item Was the “raw” data saved in addition to the preprocessed/cleaned/labeled
data (e.g., to support unanticipated future uses)?
    \begin{itemize}
        \item Yes, each QA data is accompanied with reasoning skill types and a grade level for comprehensive analysis of different benchmarks.
    \end{itemize}
    \item Is the software used to preprocess/clean/label the instances available?
    \begin{itemize}
        \item The data preprocessing and cleaning was done using Python.
    \end{itemize}
    
\end{itemize}

\subsection{Use Cases}

\begin{itemize}
    \item Has the dataset been used for any tasks already?
    \begin{itemize}
        \item Yes, we developed a baseline model of cross-modal Transformers and multiple benchmarks for icon question answering, and we trained the models on the IconQA dataset. For more details, refer to Section \ref{sec:benchmark} of the main paper.
    \end{itemize}
    \item Is there a repository that links to any or all papers or systems that
use the dataset?
    \begin{itemize}
        \item Yes, you can access the code to our model at \url{github.com/lupantech/IconQA}.
    \end{itemize}
    \item What (other) tasks could the dataset be used for?
    \begin{itemize}
        \item Currently, the dataset is intended for training visual question answering systems to access the abilities of diagram upstanding and visual reasoning. More uses could be explored in research of computer vision, natural language processing, and multimodal learning, as well as applications in smart education like tutoring systems.
    \end{itemize}
    \item Is there anything about the composition of the dataset or the way
it was collected and preprocessed/cleaned/labeled that might impact future uses?
    \begin{itemize}
        \item No.
    \end{itemize}
    
\end{itemize}

\subsection{Distribution}

\begin{itemize}
    \item Will the dataset be distributed to third parties outside of the entity (e.g., company, institution, organization) on behalf of which the dataset was created?
    \begin{itemize}
        \item The dataset is free to all under the condition that the dataset is used for non-commercial purposes only.
    \end{itemize}
    \item How will the dataset will be distributed (e.g., tarball on website, API, GitHub)?
    \begin{itemize}
        \item You can find our dataset both on the IconQA website \url{iconqa.github.io}, or the github repository \url{github.com/lupantech/IconQA}
    \end{itemize}
    \item Will the dataset be distributed under a copyright or other intellectual property (IP) license, and/or under applicable terms of use (ToU)?
    \begin{itemize}
        \item The dataset will be distributed under the CC BY-NC-SA (Attribution-NonCommercial-ShareAlike) license\footnote{\url{https://creativecommons.org/licenses/by-nc-sa/4.0}}.
    \end{itemize}
    \item Have any third parties imposed IP-based or other restrictions on the data associated with the instances?
    \begin{itemize}
        \item The source of the data instances, IXL, does not allow the data to be used commercially.
    \end{itemize}

\end{itemize}

\subsection{Maintenance}

\begin{itemize}
    \item Who is supporting/hosting/maintaining the dataset?
    \begin{itemize}
        \item The dataset is maintained by the paper's authors.
    \end{itemize}
    \item How can the owner/curator/manager of the dataset be contacted?
    \begin{itemize}
        \item The contact information of the authors can be found at the beginning of the main paper.
    \end{itemize}
    \item Is there an erratum?
    \begin{itemize}
        \item Currently, little errors have been found in the dataset. However, if errors were to be found, an erratum will be included in the repository.
    \end{itemize}
    \item Will the dataset be updated (e.g., to correct labeling errors, add new instances, delete instances)?
    \begin{itemize}
        \item If the dataset were to be updated, all versions will be available on the dataset website.
    \end{itemize}
    \item If others want to extend/augment/build on/contribute to the dataset, is there a mechanism for them to do so?
    \begin{itemize}
        \item Please contact the author through email.
    \end{itemize}

\end{itemize}

\subsection{Novelty}
IconQA presents new challenges in icon understanding and cognitive reasoning to many existing visual reasoning methods. 1) Icons feature intrinsic natures of abstract symbolism, varied styles, and ambiguous semantics, which differs from natural images significantly. 2) Since there is a lack of high-quality annotation data for icon diagrams, it restricts current mainstream data-driven visual methods to transfer smoothly to the icon domain. 3) As 107,439 questions in IconQA stem from real-world math word problems, it has made 13 different cognitive reasoning skills essential, including spatial reasoning, commonsense reasoning, estimation, and arithmetic calculation.

\subsection{Limitations and Future Work}
\label{app:limitation}
\textbf{Dataset Expansion.} As discussed in Section \ref{sec:iconqa}, IconQA focuses on colored abstract diagrams and questions of third grade and below to simplify the context scenarios and attract the community's attention on diagram understanding and visual reasoning. We would like to expand the dataset to provide greater diversity of diagram formats, grade levels, icon classes and reasoning skill types.

\textbf{Fine-grained Annotations.} IconQA benchmarks the visual question answering task in the icon domain and releases a dataset of questions, diagrams and answers. But it would be beneficial to include the object-level parsing annotations and textual explanations for each diagram and question, which facilitates future research on semantic diagram parsing and transparent visual reasoning.

\section{The Icon645 Dataset}
\label{app_icon645}

The following datasheet follows the format suggested in this paper \cite{gebru2018datasheets}.

\subsection{Dataset Statistics}
\begin{table}[ht!]
    \centering
    \small
    \caption{Statistics for the Icon645 dataset.}
    \renewcommand\tabcolsep{5.0pt}
    \begin{tabular}{lcccccc}
        \toprule	
        \textbf{Data} & \#Classes & \#Icons  & Min Size & Max Size & Colored \\ 
        \midrule	
        Icon645 & 377 & 645,687 & 64$\times$64 & 256$\times$256 &  \checkmark \\ 
        \bottomrule	
    \end{tabular}
    \vspace{0mm}
    \label{table:icon}
\end{table}

\subsection{Dataset Label}
\begin{figure}[ht!]
    \centering
    \resizebox{0.6\linewidth}{!}{
        \sffamily
{
\fbox{%
    \begin{tabular}{@{}p{\NFwidth}@{}}
    
    \multicolumn{1}{c}{\huge\bfseries Icon645 Dataset Facts}\\
    \NFrule
    \NFelement{Website}{}{\url{https://iconqa.github.io}}\\
    \NFRULE
    \NFline{Metadata}\\
    \NFrule
    \NFelement{Classes}{}{377}\\
    \NFelement{Instances}{}{645,687}\\
    \NFelement{Format}{}{.png}\\
    \NFelement{Image Sizes}{}{64$\times$ 64 - 256 $\times$ 256}\\
    \NFelement{License}{}{CC BY-NC-SA}\\
    \NFelement{Original Use Case}{}{Training visual encoder}\\
    
    \NFRule
    \NFline{Composition}\\
    \NFrule
    \NFelement{Sample or Complete}{}{Sample from \text{flaticon.com}}\\
    \NFelement{Missing Data}{}{No data is missing}\\
    
    \NFRule
    \NFline{Collection}\\
    \NFrule
    \NFelement{Sampling Strategy}{}{See the main paper}\\
    \NFelement{Author Consent}{}{None needed}\\
    
    \NFRule
    \NFline{Cleaning and Labeling}\\
    \NFrule
    \NFelement{Cleaning Done}{}{Repetitions and redundancies removed}\\
    \NFelement{Labeling Done}{}{Yes}\\
    
    \NFRule
    \NFline{Uses and Distribution}\\
    \NFrule
    \NFelement{Notable Uses}{}{Pre-training abstract icon image encoder}\\
    \NFelement{Other Uses}{}{Abstract image classification, transfer learning}\\
    \NFelement{Original Distribution}{}{Check dataset website}\\
    
    \NFRule
    \NFline{Maintenance and Evolution}\\
    \NFrule
    \NFelement{Corrections or Erratum}{}{None as of now}\\
    \NFelement{Methods to Extend}{}{Contact the author}\\
    
    \NFRULE
    \end{tabular}}
\par} 
        }
    \caption{Icon645 dataset label, created with the template from the paper \cite{bandy2021addressing}.}
    \label{fig:card_icon645}
\end{figure}

\subsection{Links}

The link to download the IconQA dataset can be found on \url{iconqa.github.io}.

\subsection{Motivation}

\begin{itemize}
    \item \ques{For what purpose was the dataset created? Was there a specific task in mind? Was there a specific gap that needed to be filled?}
    \begin{itemize}
        \item Icon645 was created for the purpose of pre-training image encoders on the icon image classification task. Presently, no other dataset provides such a large variety of abstract icons with appropriate labels.
    \end{itemize}
    \item \ques{Who created the dataset (e.g., which team, research group) and on
behalf of which entity (e.g., company, institution, organization)?}
    \begin{itemize}
        \item This dataset was created under the combined effort of multiple researchers from University of California, Los Angeles, Sun Yat-sen University, East China Normal University, and Columbia University.
    \end{itemize}
    \item \ques{Who funded the creation of the dataset?}
    \begin{itemize}
        \item The project received no funding or associated grant.
    \end{itemize}
\end{itemize}

\subsection{Composition}

\begin{itemize}
    \item \ques{What do the instances that comprise the dataset represent (e.g.,
documents, photos, people, countries)? Are there multiple types of
instances (e.g., movies, users, and ratings; people and interactions between them; nodes and edges)?}
    \begin{itemize}
        \item Each instance is a single colored icon image with size between $64 \times 64$ and $256 \times 256$ pixels.
    \end{itemize}
    \item \ques{How many instances are there in total (of each type, if appropriate)?}
    \begin{itemize}
        \item There are a total of 645,687 instances categorized into 377 classes.
    \end{itemize}
    \item \ques{Does the dataset contain all possible instances or is it a sample
(not necessarily random) of instances from a larger set?}
    \begin{itemize}
        \item The dataset is a sample of the Flaticon library. Only 377 classes of icons that satisfy our requirements outlined in the paper are included in the dataset.
    \end{itemize}
    \item \ques{What data does each instance consist of? }
    \begin{itemize}
        \item Each instance is a single icon image
    \end{itemize}
    \item \ques{Is there a label or target associated with each instance?}
    \begin{itemize}
        \item Yes, Each image is given a text label, specifying its class.
    \end{itemize}
    \item \ques{Is any information missing from individual instances?}
    \begin{itemize}
        \item No. All related information is included in the dataset.
    \end{itemize}
    \item \ques{Are there recommended data splits (e.g., training, development/validation, testing)?}
    \begin{itemize}
        \item No. The user can decide how they want to split the dataset.
    \end{itemize}
    \item \ques{Are there any errors, sources of noise, or redundancies in the
dataset?}
    \begin{itemize}
        \item No.
    \end{itemize}
    \item \ques{Is the dataset self-contained, or does it link to or otherwise rely on external resources (e.g., websites, tweets, other datasets)?}
    \begin{itemize}
        \item The dataset is self-contained. All related information is included in the dataset.
    \end{itemize}
    \item \ques{Does the dataset contain data that might be considered confidential?}
    \begin{itemize}
        \item No, the dataset does not contain anything related to any individual.
    \end{itemize}
    \item Does the dataset contain data that, if viewed directly, might be offensive, insulting, threatening, or might otherwise cause anxiety?
    \begin{itemize}
        \item No, the dataset does not contain anything offensive.
    \end{itemize}
\end{itemize}

\subsection{Collection Process}
\begin{itemize}
    \item How was the data associated with each instance acquired?
    \begin{itemize}
        \item The data is publicly available on flaticon.com. More details are included in the main paper.
    \end{itemize}
    \item What mechanisms or procedures were used to collect the data
(e.g., hardware apparatus or sensor, manual human curation, software program, software API)?
    \begin{itemize}
        \item We implemented a program to retrieve the target 377 icon classes using Python.
    \end{itemize}
    \item Over what timeframe was the data collected?
    \begin{itemize}
        \item The dataset was finally completed in March, 2021 after three months of data collection, cleaning and prepossessing.
    \end{itemize}
    \item Does the dataset relate to people?
    \begin{itemize}
        \item No, the dataset does not relate to people.
    \end{itemize}
\end{itemize}

\subsection{Preprocessing}

\begin{itemize}
    \item Was any preprocessing/cleaning/labeling of the data done (e.g.,
discretization or bucketing, tokenization, part-of-speech tagging,
SIFT feature extraction, removal of instances, processing of missing values)?
    \begin{itemize}
        \item We cropped white space from each icon diagram in Icon645 to tighten it up. Black and white icons were filtered out. Redundant instances were removed based on the metric of diagram similarity.
    \end{itemize}
    \item Was the “raw” data saved in addition to the preprocessed/cleaned/labeled
data (e.g., to support unanticipated future uses)?
    \begin{itemize}
        \item No.
    \end{itemize}
    \item Is the software used to preprocess/clean/label the instances available?
    \begin{itemize}
        \item The data preprocessing and cleaning was done using Python.
    \end{itemize}
    
\end{itemize}

\subsection{Use Cases}

\begin{itemize}
    \item Has the dataset been used for any tasks already?
    \begin{itemize}
        \item Yes, we have used the dataset to pre-train an abstract image encoder to act as the backbone network in our Patch-TRM model.
    \end{itemize}
    \item Is there a repository that links to any or all papers or systems that
use the dataset?
    \begin{itemize}
        \item Yes, you can access the code to our model at \url{github.com/lupantech/IconQA}.
    \end{itemize}
    \item What (other) tasks could the dataset be used for?
    \begin{itemize}
        \item Currently, the dataset is intended for training abstract icon image classifiers. Other possibilities could be explored in the future.
    \end{itemize}
    \item Is there anything about the composition of the dataset or the way it was collected and preprocessed/cleaned/labeled that might impact future uses?
    \begin{itemize}
        \item No.
    \end{itemize}
    
\end{itemize}

\subsection{Distribution}

\begin{itemize}
    \item Will the dataset be distributed to third parties outside of the entity (e.g., company, institution, organization) on behalf of which the dataset was created?
    \begin{itemize}
        \item The dataset is free to all under the condition that the dataset is used for non-commercial purposes only.
    \end{itemize}
    \item How will the dataset will be distributed (e.g., tarball on website, API, GitHub)?
    \begin{itemize}
        \item The dataset will be accessible on \url{github.com/lupantech/IconQA}
    \end{itemize}
    \item Will the dataset be distributed under a copyright or other intellectual property (IP) license, and/or under applicable terms of use (ToU)?
    \begin{itemize}
        \item The dataset will be distributed under the CC BY-NC-SA (Attribution-NonCommercial-ShareAlike) license\footnote{\url{https://creativecommons.org/licenses/by-nc-sa/4.0}}.
    \end{itemize}
    \item Have any third parties imposed IP-based or other restrictions on the data associated with the instances?
    \begin{itemize}
        \item The source of the data instances, Flaticon, does not allow the data to be used commercially.
    \end{itemize}

\end{itemize}

\subsection{Maintenance}

\begin{itemize}
    \item Who is supporting/hosting/maintaining the dataset?
    \begin{itemize}
        \item The dataset is maintained by the paper's authors.
    \end{itemize}
    \item How can the owner/curator/manager of the dataset be contacted?
    \begin{itemize}
        \item The contact information of the authors can be found at the beginning of the main paper.
    \end{itemize}
    \item Is there an erratum?
    \begin{itemize}
        \item Currently, no error has been found in the dataset. However, if errors were to be found, an erratum will be included in the repository.
    \end{itemize}
    \item Will the dataset be updated (e.g., to correct labeling errors, add new instances, delete instances)?
    \begin{itemize}
        \item If the dataset were to be updated, all versions will be available on the dataset website.
    \end{itemize}
    \item If others want to extend/augment/build on/contribute to the dataset, is there a mechanism for them to do so?
    \begin{itemize}
        \item Please contact the author through email.
    \end{itemize}

\end{itemize}

\section{Details of Baseline Patch-TRM}
\label{app_patchtrm}
We develop a patch cross-modal Transformer model (Patch-TRM) as a strong baseline for the IconQA task as illustrated in Figure \ref{fig:model}. We will introduce the details of Patch-TRM as follows.


\subsection{Diagram Encoder}
Similar to natural images in most VQA datasets, abstract diagrams also have rich visual and semantic information that is critical to answering questions. Current dominant VQA methods \cite{antol2015vqa, Anderson2017up,Kim2018,gao2019dynamic,yu2019mcan,jiang2020defense,agarwal2020towards} either extract high-level visual representations from a pre-trained ResNet backbone network \cite{he2016deep} in a top-down fashion, or apply a bottom-up mechanism to extract semantic representations via a object detector, such as a model based on Faster R-CNN \cite{ren2015faster}. However, these methods depend heavily on the backbone network, which is pre-trained on natural images. When processing diagrams in IconQA, they are likely to fail to extract meaningful representations or reasonable object proposals. Inspired by the early progress in using hierarchical scene layout to parse images \cite{li2010object, zhu2015reconfigurable,wang2015learning} and the recent advances in Transformer-based image encoding \cite{lu2019vilbert, li2019visualbert,wonjae2021an}, we develop a method that splits diagrams into hierarchical patch sequences from a pyramid structure and learns their visual representations using a visual Transformer.

As diagrams in IconQA have more varied aspect ratios than natural images, we add blank paddings at the bottom or on the right side of the images to ensure that they are square-shaped. Each padded diagram is then cropped into a set of patch sequences with different scales. The padding operation and the hierarchical scene layout can facilitate extracting complete objects that retain specific semantics. Let $p = [p_1, p_2, \dots , p_n]$ denote the patch sequence in the splitting order from the original diagram. From each patch sequence, we extract the visual features using a ResNet model and represent the features as $f_p = [f_{p_1}, f_{p_2}, \dots , f_{p_n}]$. The representation for each patch, $f_{p_i}$, is then summed up with its positional embedding with respect to its sequencial index $i$. Finally, the updated visual patch embeddings pass through a standard multi-layer Transformer \cite{vaswani2017attention} to learn high-level visual representations $h_p = [h_{\mathtt{[CLS]}}, h_{p_1}, h_{p_2}, \dots , h_{p_n}]$. Here, the trainable token $\mathtt{[CLS]}$, which is added to the Transformer inputs, learns the global meaning of these sequences. As mentioned before, it is not feasible to use existing pre-trained ResNet to process abstract diagrams due to a lack of similar resources for pre-training. So we pre-train the ResNet on icon classification with the icon dataset we compiled (Section \ref{sec:Icon645}). More details of the pre-training task are discussed in Section \ref{sec:icon_class}.

\subsection{Language Encoder}
Questions in IconQA have a wide distribution of question lengths, so we follow the recent approaches \cite{vaswani2017attention,jiao2019tinybert,turc2019well,li2019visualbert, lu2019vilbert} that apply the BERT model \cite{devlin2018bert} to embed question texts, rather than using traditional LSTM \cite{hochreiter1997long} or GRU \cite{cho2014learning} for long sequence encoding. Given the question $w_0, w_1, \dots, w_t$, the input is formatted as $[\mathtt{[CLS]}, w_0, w_1, \dots, w_t]$. We use the WordPiece \cite{schuster2012japanese} subword tokenizer and the resulting sequence is padded to the maximum length. Similar to other methods that use BERT as sentence encoders, we consider the output corresponding to the first token $\mathtt{[CLS]}$ as the embedding of the entire question, noted as $h_q$.

\subsection{Answer Reasoning}
Given the image patch representation $h_p \in \mathcal{R}^{n \times k}$, and question embedding $h_q \in \mathcal{R}^{k}$, where $n$ denotes the number of diagram patches and $k$ denotes the learned embedding size of the patches, we apply a cross-modal attention to learn their joint representation:
\begin{align}
a &=\operatorname{softmax}\left(W_{p} h_p \circ W_{q} h_q\right) ,\\ 
h_v &=\sum_{i}^{n} a(i) \times h_{p_i} ,
\end{align}
where $W_{p}$ and $W_{q}$ are learnable mapping parameters, and $\circ$ is the element-wise product operator. The joint representation $h_v$ is calculated as the weighted sum over all diagram patches.

Before predicating the answer, multiple choice candidates need to be encoded. Taking the \textit{multi-image-choice} task as an example, each image choice is encoded as the output of the last pooling layer of the pre-trained ResNet. The encoded image choice is denoted as $h_c \in \mathcal{R}^{m\times k} $, where $m$ is the number of the candidates. The choice embeddings are concatenated with the diagram-question representation, and then the resulted embeddings are fed to a classifier over the candidates:
\begin{align}
p_{ans}=\operatorname{softmax}\left(W_{a}\left([h_{v};h_{c}]\right)+b_{a}\right), 
\end{align}
where $W_{a}$ and $b_{a}$ are classifier parameters, and $p_{\text {ans}}$ is the probability of the predicated answer choice. 

Similarly, in the \textit{multi-text-choice} sub-task, the answer is predicated over text choices, except that each text choice is embedded with LSTM layers first. 
We formulate the \textit{filling-in-the-blank} sub-task as a multi-class classification problem from all possible answers in the training data, as most VQA works do. After generating the joint encoding for the input diagram and question, a linear classifier is trained to predict the final answer.

\section{User Study}
\label{app_user_study}

\subsection{Crowd Sourcing Method}
Using Amazon Mechanical Turk (AMT), we ask people to provide answers to the questions in the test set along with their age group. We also strongly encourage parents who have young children to let their children complete the questionnaires, as their answers give us insights to how the designed audience of these questions perform. The test set is split into batches of 20 questions, which we call a task, with each task assigned to 3 crowd workers on AMT. This amounts to a total of 64,467 effective test set answers.

\subsection{Quality Assurance}

To ensure the truthfulness of the age information, we ask the participants to select their age at both the beginning and the end of the questionnaire, with the age choices appearing in 2 different orders. \par
\begin{figure*}[t!]
\centering 
    \includegraphics[width= 0.98\linewidth]{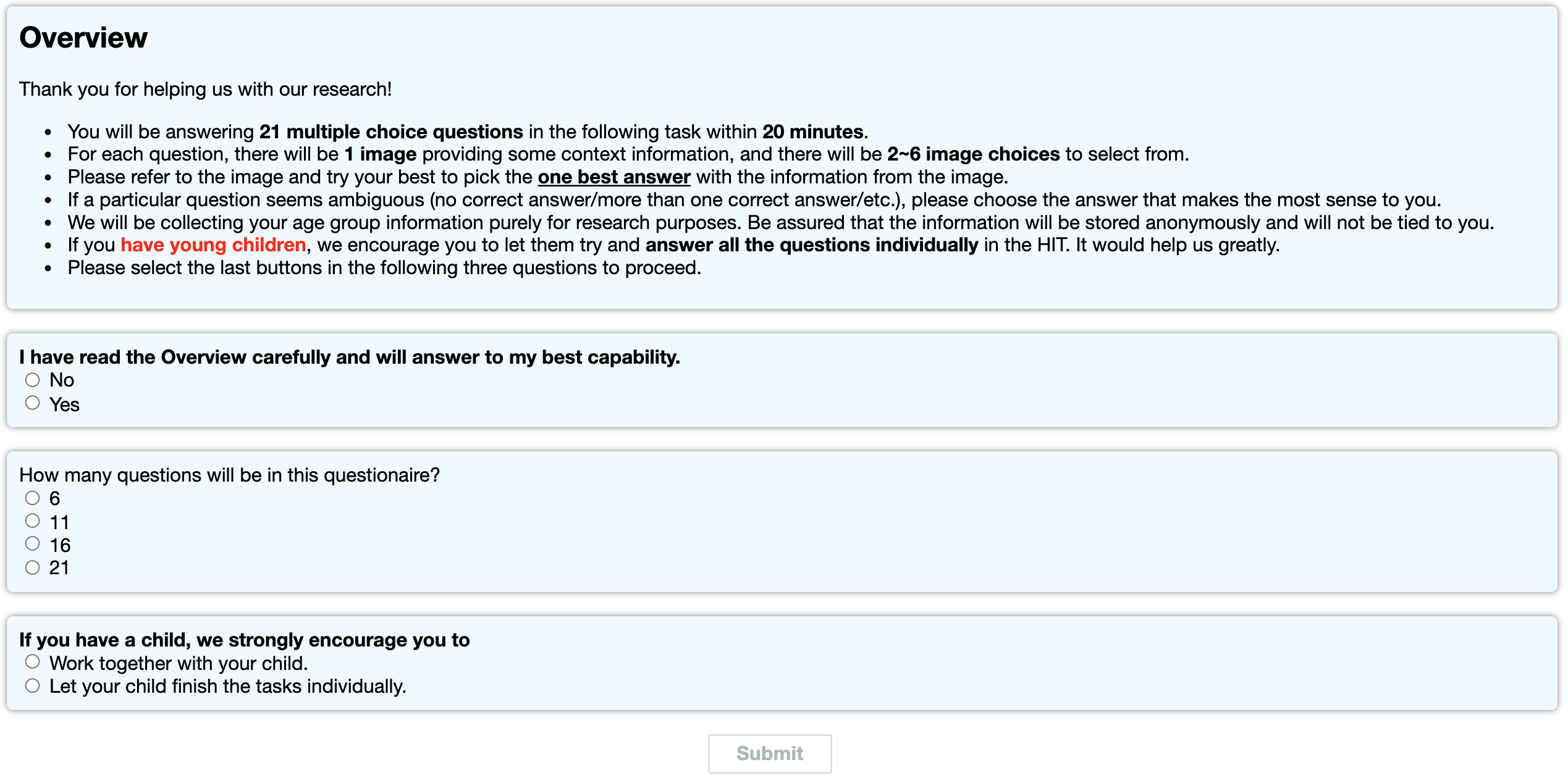}
    \caption{AMT instructions for the user study.}
    \label{fig:amt_instruction}
\end{figure*}
\begin{figure*}[t!]
\centering 
    \includegraphics[width= 0.98\linewidth]{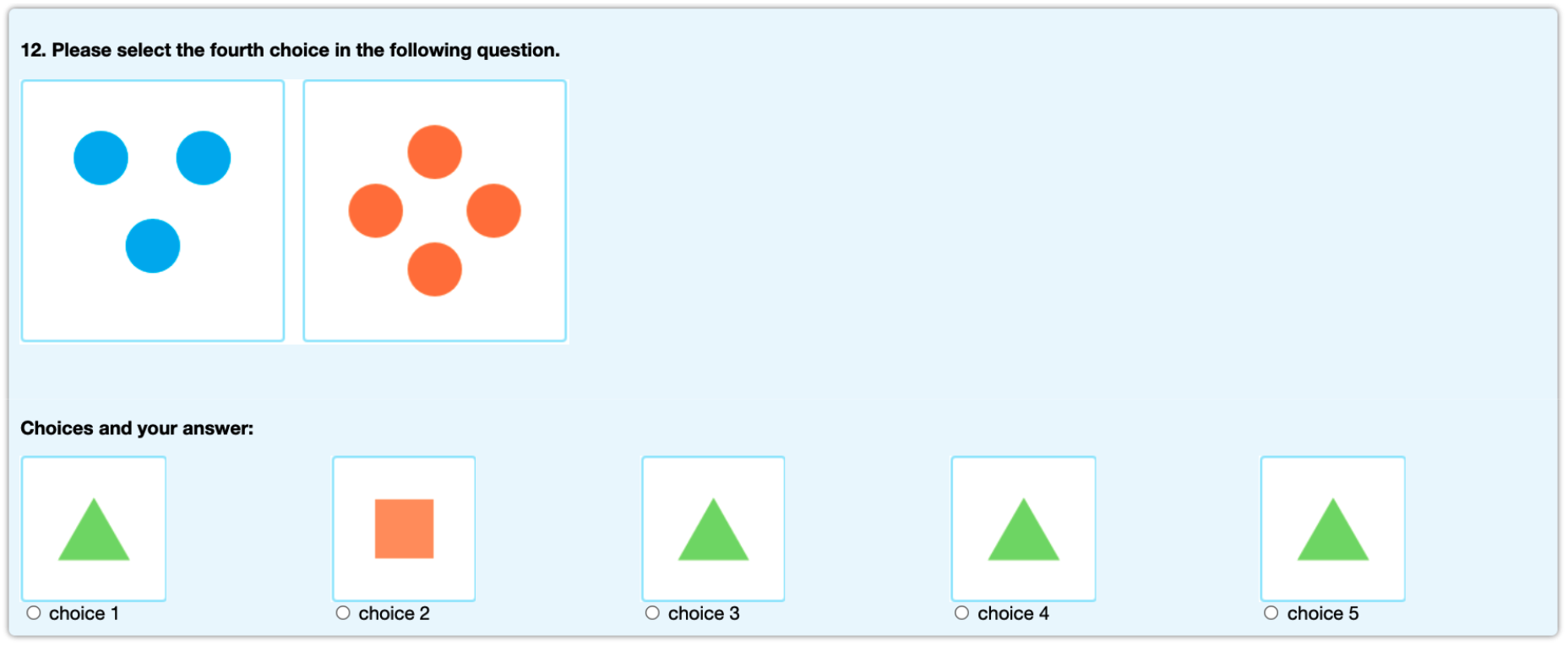}
    \includegraphics[width= 0.98\linewidth]{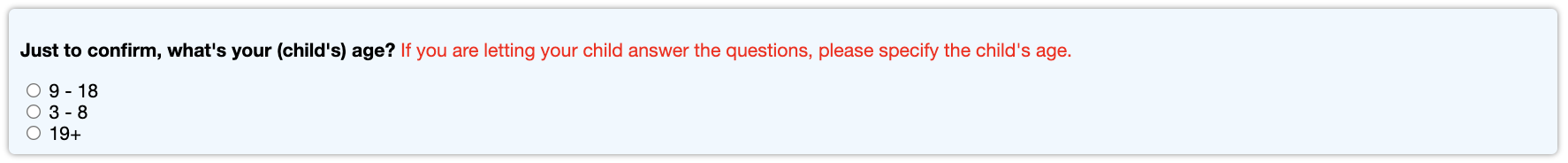}
    \caption{AMT attention check questions.}
    \label{fig:amt_check_ques}
\end{figure*}
\begin{figure*}[t!]
\centering 
    \includegraphics[width= 0.98\linewidth]{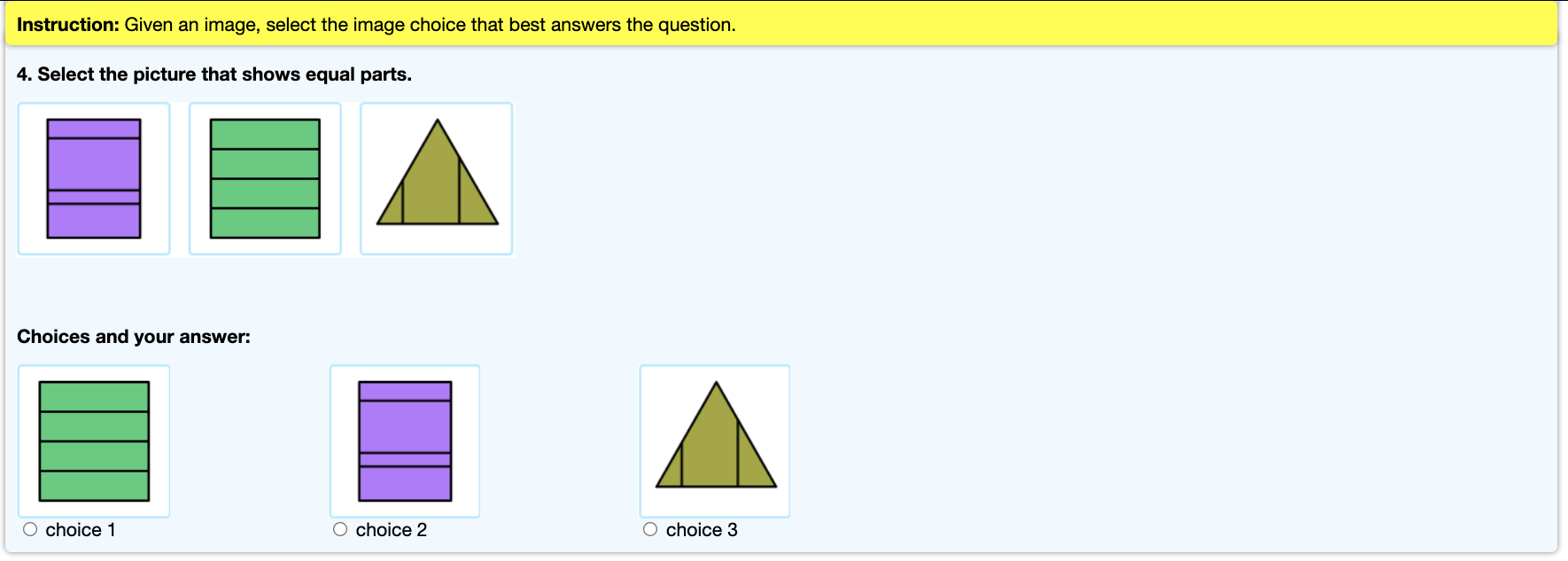}
    \caption{An AMT question example for the \textit{multi-image-choice} sub-task.}
    \label{fig:amt_example_choose_img}
\end{figure*}
\begin{figure*}[t!]
\centering 
    \includegraphics[width= 0.98\linewidth]{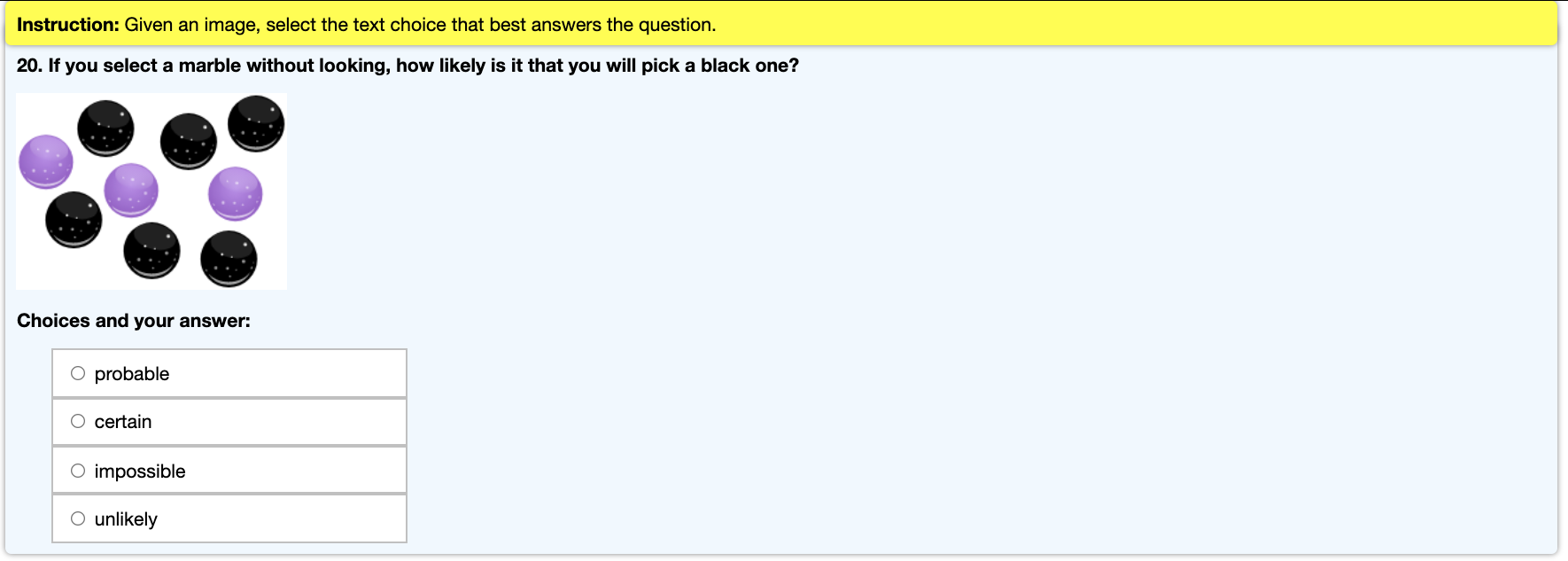}
    \caption{An AMT question example for the \textit{multi-text-choice} sub-task.}
    \label{fig:amt_example_choose_txt}
\end{figure*}
\begin{figure*}[t!]
\centering 
    \includegraphics[width= 0.98\linewidth]{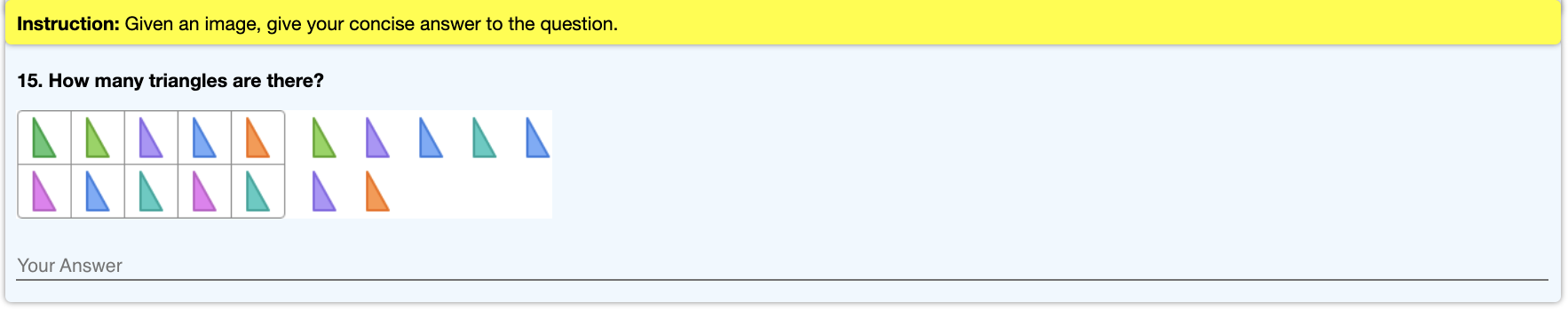}
    \caption{An AMT question  example for the \textit{filling-in-the-blank} sub-task.}
    \label{fig:amt_example_fill_in_blank}
\end{figure*}

To ensure the quality of the answers, we include 4 attention check questions: 3 of which are about the instructions, making sure that the participants read the instructions carefully. We also add an extra fake question in the middle for each \textit{choosing an image choice} and \textit{choosing a text choice} task, instructing them to choose the fourth choice despite what the choices are. Figure \ref{fig:amt_instruction} shows the instructions and the first three attention check questions. Figure \ref{fig:amt_check_ques} shows the fake question along with the age confirmation. Figure \ref{fig:amt_example_choose_img}, \ref{fig:amt_example_choose_txt}, and \ref{fig:amt_example_fill_in_blank} are example questions for three sub-tasks respectively. We also make sure that the workers answering our tasks have a history HIT approval rate of at least 95\% and a previous approval count of 1,000. 

In summary, for each Human Intelligence Task (HIT) on AMT, we have 2 age questions, 4 attention check questions, and 20 real questions from the IconQA test set. Among the 64,467 test answers, we filter out 1) the questionnaires that do not pass the 4 attention check questions, 2) the questionnaires that do not answer consistently for the two age-related questions, 3) the questionnaires that are finished unreasonably slowly/quickly. After filtering, we have 54,896 effective question answers, which we believe is a decently large sample for the human performance study.\par

\subsection{Worker Compensation}
\label{app:compensation}

For each batch of \textit{multiple choice} questions, we provide a monetary compensation of 10 US cents. For each batch of \textit{filling-in-the-blank} questions, we provide a compensation of 20 cents due to the increased difficulty. We decide upon these numbers after a few timed test trials run by ourselves. we find that these numbers enable the workers to acquire 6 USD per hour, an above average hourly wage on the AMT platform \cite{hara2018data}. The total spending in the end sums up to 452.52 USD.

\section{Experiments} 
\label{app_exp}

\subsection{Experimental Details}
\label{app_experiment_detail}

We use the same learning parameters set in Top-Down \cite{Anderson2017up} when evaluating the eight benchmarks listed in Section \ref{sec:benchmark} and our developed baseline \text{Patch-TRM}. Some crucial parameters used in our model are clarified below. 

\textbf{Our Baseline Model}. For our baseline Patch-TRM, each diagram is split four times by varied scales, resulting in 79 (1+4+9+16+49) patches totally. After resizing them to to 224$\times$224, patch visual features are extracted from the last pooling layer, resulting in a 2048-d feature vector. The ResNet network used to embed the patches is pre-trained on the icon classification task as discussed in Section \ref{sec:icon_class}. The patch Transformer has one layer of Transformer block with four attention heads and outputs embeddings with a hidden state size of 768. A small pre-trained BERT model \cite{turc2019well} is used to encode the question text in the language encoder. 

\textbf{Attention models.} For Top-Down, the attention-based baselines use 7$\times$7$\times$2048-d features from the last convolution layer. For BAN \cite{Kim2018}, DFAF \cite{gao2019dynamic}, and MCAN \cite{yu2019mcan}, image features of dimension 2,048 are extracted from Faster R-CNN \cite{ren2015faster}. Question words in these attention models are encoded into features of dimension 1,024 by GRU \cite{cho2014learning}. And the visual and textual features are then embedded into 1,024 dimensions with the corresponding attention mechanisms and fusion methods reported in original works. 

\textbf{Transformer models.} For ViLBERT \cite{lu2019vilbert} and UNITER \cite{chen2020uniter}, we use Faster R-CNN \cite{ren2015faster} to extract 36 proposal regions as the visual inputs. Both ViL \cite{wonjae2021an} and ViLT \cite{pmlr-v139-kim21k} use ViT-B/32 pre-trained on ImageNet to encode the image emebeddings. The hidden size is set as 768, the layer depth is 32, and the input image is sliced into patches with a size of 32. For ViL, we use two dependent Transformers to embed the question and image respectively.

\subsection{Human Performance}
The detailed results for human performance in the IconQA task are shown in Table \ref{table:human}.
\begin{table*}[ht!]
\centering
\scriptsize
\renewcommand\tabcolsep{2.5pt}
\begin{tabular}{{l}*{16}{c}}
    \toprule
    \multirow{3}{*}{} &
    \multicolumn{3}{c}{\textbf{Sub-tasks (3)}} & \multicolumn{13}{c}{\textbf{Reasoning skills (13)}} 	\\
    \cmidrule(lr){2-4} \cmidrule{5-17} 
    \textbf{Method} & Img. & Txt. & Blank 
    & Geo. & Cou. & Com. & Spa. & Sce. & Pat. & Tim. & Fra. & Est. & Alg. & Mea. & Sen. & Pro. \\
    \midrule
    Human & 95.69 & 93.91 & 93.56 & 94.63 & 97.63 & 94.41 & 93.31 & 92.73 & 95.66 & 97.94 & 97.45 & 87.51 & 96.29 & 86.55 & 97.06 & 85.67 \\
    Human (3-8) & 94.58 & 89.51 & 89.61 & 93.02 & 96.20 & 91.28 & 91.24 & 90.45 & 95.76 & 95.32 & 97.54 & 78.86 & 95.33 & 78.57 & 93.92 & 74.76 \\
    Human (9-18) & 94.63 & 90.97 & 93.71 & 93.28 & 97.04 & 93.46 & 91.47 & 90.92 & 94.55 & 97.59 & 96.77 & 86.79 & 95.83 & 86.60 & 96.51 & 80.56 \\
    Human (19+) & 97.34 & 95.83 & 94.22 & 96.27 & 98.44 & 96.17 & 96.31 & 95.85 & 96.34 & 98.96 & 97.95 & 89.59 & 96.84 & 88.00 & 98.49 & 90.82 \\
    \bottomrule	
\end{tabular}
\caption{Human performance in the IconQA task.}
\label{table:human}
\end{table*}

\subsection{Quantitative Analysis}

\begin{figure*}[th!]
    \centering 
    \includegraphics[width= 0.92\linewidth]{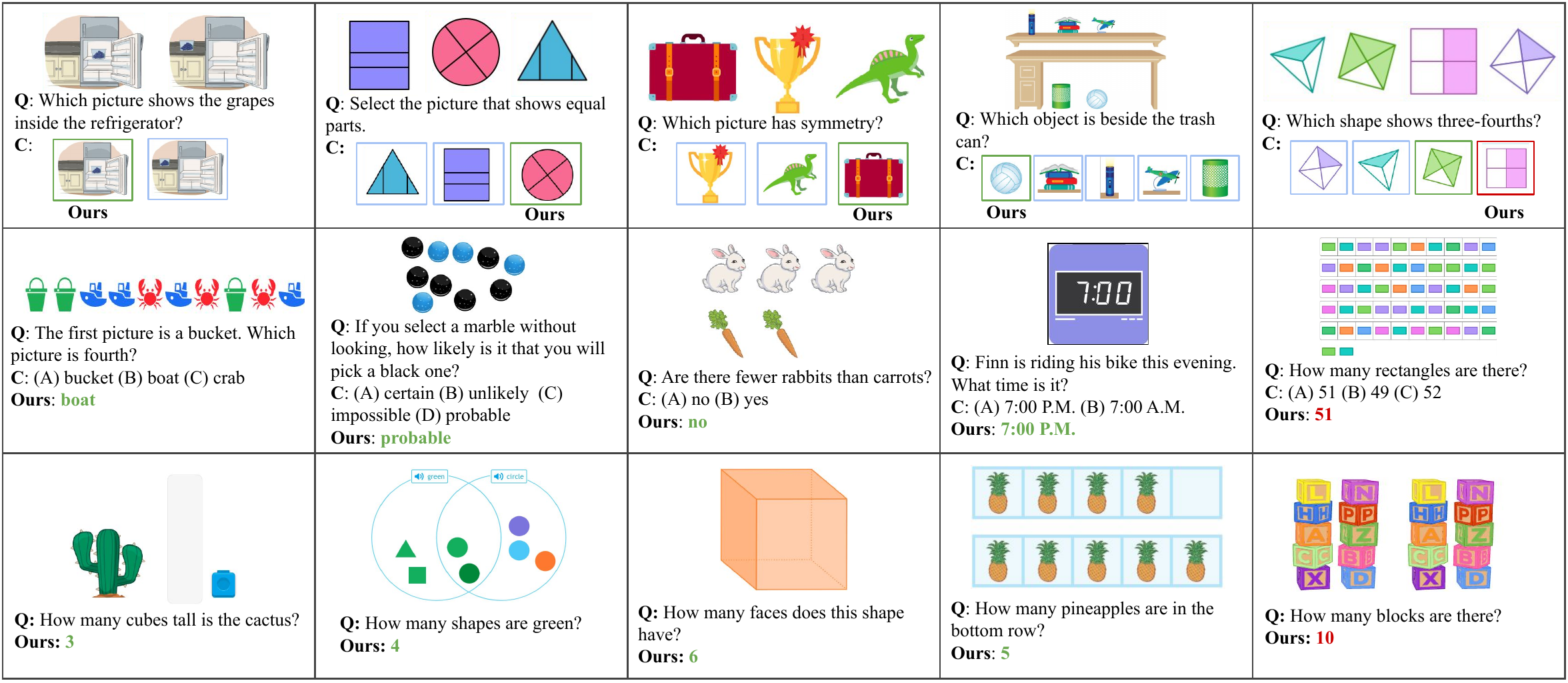}
    \vspace{0mm}
    \caption{Result examples predicted by our \text{Patch-TRM} model in the IconQA test set. \textbf{Top}:  \textit{Multi-image-choice} sub-task. \textbf{Middle}: \textit{ Multi-text-choice} sub-task. \textbf{Bottom}: \textit{Filling-in-the-blank} sub-task. Correctly predicted answers are highlighted by green, while wrong ones are highlighted by red.}
    \label{fig:result_examples}
\end{figure*}

Figure \ref{fig:result_examples} presents five examples from the IconQA test set predicted by our  \text{Patch-TRM} baseline for each sub-task. Although \text{Patch-TRM} achieves promising results for most problems in IconQA, it still fails to address some complicated cases. For example, it encounters difficulties in identifying dense objects and making multi-hop reasoning.

\end{document}